\documentclass[11pt,letterpaper]{article}

\usepackage[margin=1in]{geometry}
\usepackage[authoryear,round]{natbib}
\setcitestyle{citesep={;},aysep={,},yysep={;}}
\usepackage{microtype}
\usepackage{xeCJK}
\setCJKsansfont{FandolHei-Regular.otf}[BoldFont=FandolHei-Bold.otf]
\setCJKmonofont{FandolFang-Regular.otf}

\usepackage{amsmath,amsfonts,bm}

\def\eqref#1{equation~\ref{#1}}

\def\1{\bm{1}}

\DeclareMathAlphabet{\mathsfit}{\encodingdefault}{\sfdefault}{m}{sl}
\SetMathAlphabet{\mathsfit}{bold}{\encodingdefault}{\sfdefault}{bx}{n}

\usepackage{hyperref}
\usepackage{url}
\usepackage{graphicx}
\usepackage{adjustbox}
\usepackage{booktabs}
\usepackage{placeins}
\usepackage[table]{xcolor}
\usepackage{multirow}
\usepackage{enumitem}
\usepackage{caption}
\title{PT$^3$: Prototype-Guided Self-Supervised Pretraining for Generalizable mmWave Human Point Cloud Representations}
\author{%
  Ziyang Liu \quad Kieren Yu \quad Xiao Yang \quad Xin Yang \quad Kaishun Wu\thanks{Corresponding author.}\\[0.5em]
  \small The Hong Kong University of Science and Technology (Guangzhou)
}
\date{}
\hypersetup{
  pdftitle={PT3: Prototype-Guided Self-Supervised Pretraining for Generalizable mmWave Human Point Cloud Representations},
  pdfauthor={Ziyang Liu, Kieren Yu, Xiao Yang, Xin Yang, Kaishun Wu},
  pdfsubject={Self-supervised learning for mmWave human point clouds},
  hidelinks,
  bookmarksnumbered=true
}

\begin{document}

\maketitle
\vspace{-2\baselineskip}

\begin{abstract}
Multi-source millimeter-wave (mmWave) observations offer an opportunity to learn human representations reusable across sensing tasks.
However, sparse detections and multipath effects make these observations unstable, challenging existing point cloud pretraining methods.
We propose PT$^3$, a prototype-guided self-supervised pretraining framework for mmWave human point clouds.
PT$^3$ represents local states in spatial regions as sparse compositions of shared prototypes and learns their temporal evolution through autoregressive prediction.
Observation consistency further stabilizes representations under radar perturbations.
After pretraining on unlabeled sequences from multiple sources, the same frozen encoder supports action classification, dense point generation, joint pose estimation, and future motion prediction on datasets not used for pretraining.
Across these tasks, PT$^3$ outperforms the compared pretraining baselines and achieves competitive or superior performance to task-specific models.
\end{abstract}
\section{Introduction}
\label{sec:introduction}



In the era of ubiquitous computing, applications such as smart homes, rehabilitation, human--computer interaction, and continuous health monitoring increasingly rely on contactless human sensing~\citep{yang2023mmfi,ding2024milliflow,liu2026cardiommflow}.
To enable these applications, capabilities such as action classification, dense point generation, joint pose estimation, and future motion prediction are essential.
Millimeter-wave (mmWave) radar is particularly attractive for this purpose and attract many researchers attention, as it preserves privacy by avoiding appearance images and remains robust to lighting conditions~\citep{yang2023mmfi,liu2026mmjepaecg}.



With the growing availability of mmWave human sensing datasets such as MM-Fi and M4Human~\cite{yang2023mmfi,fan2026m4human}, mmwave radar point clouds of human observations collected across devices and conditions continue to accumulate.
However, most existing methods are designed and trained for individual tasks and datasets with task-specific labels~\citep{singh2019radhar,chen2022mmbody,fan2026mmpred}, while recent radar pretraining efforts either target limited downstream task or rely on paired modalities~\citep{liang2026wave2body,lai2026radarllm}.
Building on these efforts, a promising direction is to learn shared representations of human structure and motion across sensing tasks. This motivates us to develop a unified pretraining framework that learns generalizable human representations from multiple radar datasets.

A natural starting point is general point cloud self-supervised pretraining, which learns representations through masked point modeling or autoregressive point patch generation~\citep{pang2022pointmae,yu2022pointbert,chen2023pointgpt}.
However, these objectives do not transfer directly to mmWave human point clouds, which raises three research questions:
\begin{itemize}[leftmargin=*, itemsep=2pt, topsep=3pt, parsep=0pt, partopsep=0pt]
\item \textbf{RQ1:} Why do point-level objectives transfer poorly to mmWave human point clouds?
\item \textbf{RQ2:} What representation can capture human states beyond unstable detections?
\item \textbf{RQ3:} What supervision can train such a representation across heterogeneous sources at scale?
\end{itemize}

\begin{figure}[t]
    \centering
    \adjincludegraphics[trim={0 {0.13\height} 0 0}, clip, width=0.96\linewidth]{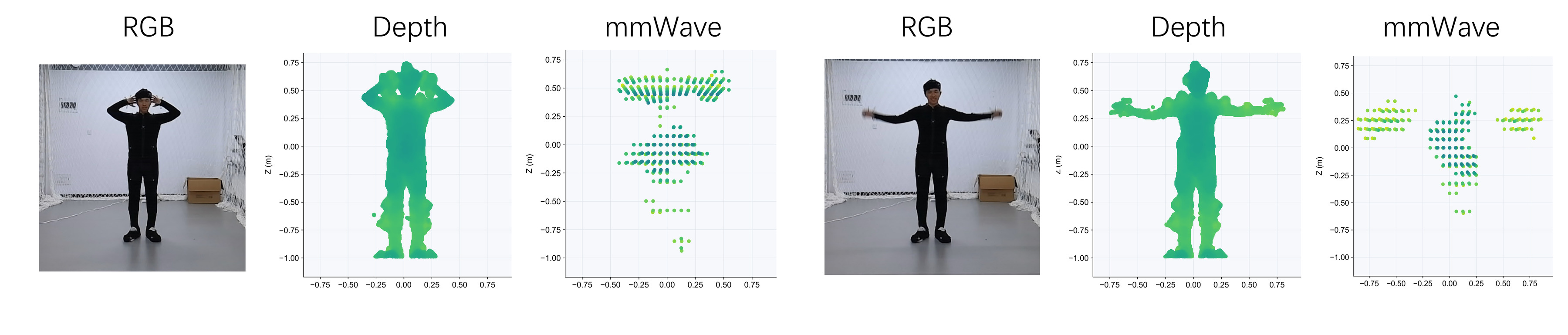}
    \caption{Comparison of RGB, depth points and sparse mmWave observations.}
    \label{fig:compare}
\end{figure}

\textbf{For RQ1.}
These objectives are designed for densely sampled point clouds, where point-level targets directly reflect surface geometry.
mmWave human point clouds violate this assumption, since their points are radar detections rather than surface samples (Fig.~\ref{fig:compare}).
CFAR detection retains only strong reflections, which depend on body orientation and viewing angle.
Multipath propagation introduces ghost points, while limited angular resolution merges nearby returns.
The resulting point sets are sparse, incomplete, and inconsistent across devices~\citep{fan2024diffusion,liang2026wave2body}.
A change in a point may thus stem from the sensing process rather than from human motion.
This suggests that the learning target should be the underlying human state rather than the detections.

\textbf{For RQ2.}
The human state is not directly observable and must be inferred from unstable detections.
Since detections lack a stable correspondence to body parts, the representation cannot be anchored to joints or anatomical parts.
It should instead be defined over a consistent spatial reference, so that states remain comparable across time and devices.
Local human states are diverse, yet they share recurring patterns across regions and time.
A representation should therefore capture these shared patterns while preserving the differences between local states.
We therefore hypothesize that local human states can be described as compositions of a shared set of latent elements.

\textbf{For RQ3.}
Learning these compositions requires supervision, yet human labels do not scale across heterogeneous sources.
Such labels require dedicated capture systems and differ in definition across datasets.
This motivates self-supervised learning that can scale with unlabeled radar datasets. 
Building on the hypothesis in RQ2, 
we aim to learn a shared set of latent elements and their compositions directly from radar observations. 
This suggets that the training objectives can be formulated around the consistency and temporal evolution of local state representations.

Following these insights, we propose PT$^3$, a prototype-guided self-supervised learning (SSL) frame for pretraining mmWave human point clouds encoder.
PT$^3$ decomposes each observation into body-agnostic spatial regions.
All regions share a single prototype bank, and each region expresses its local state as a weighted combination of a few prototypes.
Together, these regional states compose the human state, turning unstable detections into a structured representation.
Pretraining then forecasts how these compositions evolve and keeps them consistent under point dropout and spurious echoes.
Our main contributions are summarized as follows.
\begin{itemize}[leftmargin=*, itemsep=2pt, topsep=3pt, parsep=0pt, partopsep=0pt]
\item \textbf{Comprehensive SSL pretraining for mmWave human point clouds.}
To our knowledge, we present one of the most comprehensive studies of SSL pretraining for mmWave human point clouds, covering multiple data sources, devices, frequency bands, and downstream human sensing tasks.

\item \textbf{Prototype-evolution-driven SSL pretraining.}
Our model represents human states as sparse compositions of shared prototypes and learns their temporal evolution through autoregression.
Observation consistency further reduces the influence of sensing variations, enabling shared state representations to be learned jointly from heterogeneous sources.

\item \textbf{Cross-task and cross-source evaluation of representation reuse.}
We evaluate the frozen PT$^3$ encoder on datasets unseen during pretraining, covering action classification, dense point generation, joint pose estimation, and future motion prediction.
Under cross-scene, cross-subject, and cross-environment protocols, it outperforms the compared pretraining baselines
and achieves competitive or superior performance to task-specific models.
\end{itemize}





\section{Related Work}
\label{sec:related-work}

\textbf{Tasks and Datasets in mmWave Human Sensing.}
mmWave human sensing has progressed from activity and gesture recognition to pose estimation, body reconstruction, and motion modeling~\citep{singh2019radhar,palipana2021pantomime,chen2022mmbody,ding2024milliflow}.
Datasets such as MM-Fi and M4Human now cover more subjects, scenes, and tasks~\citep{yang2023mmfi,fan2026m4human}.

\textbf{General Point Cloud Pretraining and Representation Learning.}
General point cloud pretraining offers several approaches to representation learning.
Point-BERT, Point-MAE, and Point-M2AE use masked modeling to learn local and global point cloud structure~\citep{yu2022pointbert,pang2022pointmae,zhang2022pointm2ae}.
PointGPT models ordered point patches through autoregressive generation~\citep{chen2023pointgpt}.
Sonata learns general 3D representations through self-distillation on large-scale data~\citep{wu2025sonata}.
Together, these methods provide representative baselines for our evaluation of mmWave human point cloud pretraining.

\textbf{Representation Learning for mmWave Human Sensing.}
Recent work has used pretraining and cross-modal alignment to learn radar representations for human sensing.
Wave2Body pretrains an mmWave point cloud encoder with self-supervision and lable alignment for pose estimation~\citep{liang2026wave2body}.
RadarLLM converts radar point cloud sequences into semantic tokens and aligns them with a LLM~\citep{lai2026radarllm}.
Other studies explore masked modeling, cross-modal supervision, and wireless signal pretraining for human sensing~\citep{wei2026maepose,zhang2026mmmind,gong2026wiswiss}.
However, unified self-supervised pretraining across mmWave human point cloud sources and devices, and broad evaluation of a shared encoder across downstream tasks remain underexplored.

\section{Methodology}
\textbf{Problem Formulation.}
Detections in mmWave human point clouds are sparse and vary with observation conditions.
Stable correspondences between detections and body parts are therefore difficult to establish.
This makes point-level correspondence an unreliable basis for a representation shared across tasks.
We instead model the relatively stable local human states behind the detections.

\textbf{Modeling Hypothesis.}
We hypothesize that local human states can be described by combinations of reusable latent elements.
These elements are shared across spatial regions and time.
Changes in their mixture weights describe the evolution of local states.
Learning these combinations and their temporal evolution from continuous mmWave observations may yield reusable human representations that can support a range of downstream sensing tasks.

\begin{figure}[t]
    \centering
    \includegraphics[width=0.98\linewidth]{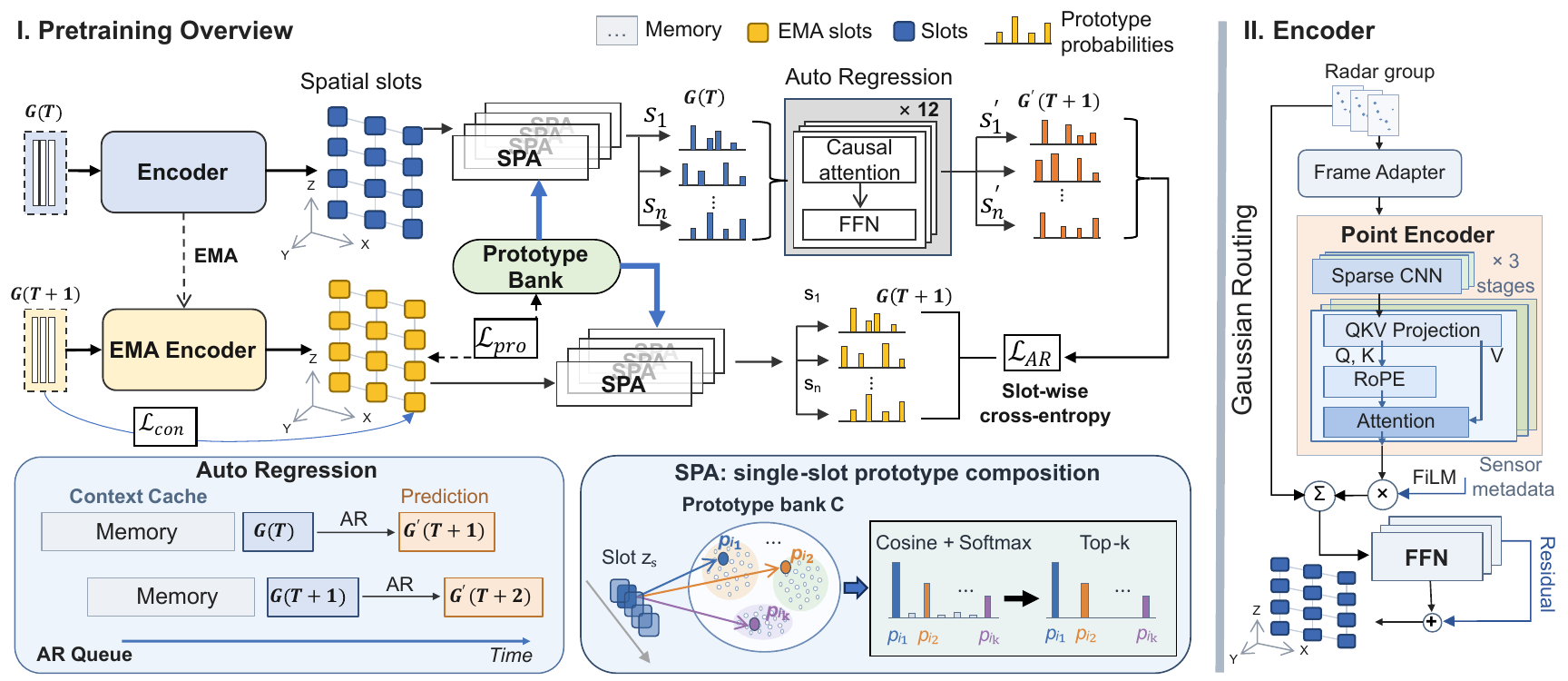}
    \caption{Overview of PT$^3$. Left: self-supervised learning of regional prototype compositions and their temporal evolution. Right: radar point cloud encoding into spatial slots shared by pretraining and downstream tasks.}
    \label{fig:overview}
\end{figure}

Let $\mathcal{X}=\{X_t\}_{t=1}^{T}$ be a mmWave point cloud sequence of short observation groups.
Each group contains detection coordinates and available attributes such as Doppler and reflection intensity.
A shared encoder $E_{\theta}$ maps each group to a continuous representation over fixed spatial regions:
\begin{equation}
    Z_t = E_{\theta}(X_t)
    \in \mathbb{R}^{S\times d},
\end{equation}
Here, $S$ is the number of regions, and $d$ is the feature dimension.
The encoder comprises observation encoding and spatial readout (Section~\ref{sec:spatial_repr}).
We use a shared prototype bank to represent each regional state as a sparse composition (Section~\ref{sec:compositional_motion}).
Three self-supervised objectives learn these compositions and their temporal evolution (Section~\ref{sec:training_objectives}).
The pretrained encoder is used for multiple downstream tasks.
Figure~\ref{fig:overview} shows the overall framework of PT$^3$ model design.

\subsection{From Radar Points to Spatial Slots}
\label{sec:spatial_repr}
We map each observation group to a fixed number of regional representations in two stages.
Observation encoding first compresses variable radar detections into representative features.
Spatial readout then aggregates them into regions with consistent indices across groups.

\textbf{Observation Encoding.}
We first resample observations to a common frame rate across sources.
We combine observations at $F$ target frame positions into $X_t$ and retain each point's original frame identity.
Let $N_t$ be the number of valid points in $X_t$.
Each point $\mathbf{x}_{t,i}\in\mathbb{R}^{5}$ contains 3D coordinates, Doppler, and reflection intensity.
Separate masks distinguish missing attributes.

Radar devices produce widely varying numbers of detections per frame.
We therefore include a lightweight Frame Adapter in the shared observation encoder.
It compresses each frame into a bounded set of representative features so that different devices can share one encoder.
The adapter groups detections by spatial location and selects one observed point as the anchor of each group.
It projects measurement attributes and offsets from the anchor into local features.
Learned attention weights aggregate these features into one representative per group.
Each representative retains the anchor's measured 3D position and summarizes local measurements.

We merge representatives from all frames and pass them through a point encoder.
The encoder extracts local features, aggregates them hierarchically, and projects them to $d$ dimensions.
Radar devices differ in hardware settings such as bandwidth and antenna layout.
Their fields of view (FoV) and spatial resolutions therefore differ, even for the same physical scene.
We use FiLM to condition the encoder on FoV bounds, spatial resolution, and validity indicators.
Weak multiplicative modulation of the projected features gives
$H_t=[\mathbf{h}_{t,1},\ldots,\mathbf{h}_{t,M_t}]^\top
\in\mathbb{R}^{M_t\times d}$,
where $M_t$ is the number of final representatives.

\textbf{Spatial Routing.}
Let $\mathbf{h}_{t,m}\in\mathbb{R}^d$ denote row $m$ of $H_t$, with spatial coordinate $\mathbf{r}_{t,m}\in\mathbb{R}^3$.
Fixed anchors aggregate these features into $S$ spatial slots.
The anchors $\{\mathbf{a}_s\}_{s=1}^{S}$ form a regular grid in the $X$--$Z$ plane.
They provide consistent spatial indices across observation groups without point-level correspondence between frames.
Nearby radar returns can jointly describe a local human state.
We therefore use overlapping soft routing to combine nearby representatives in each region.
A null slot absorbs representatives far from all anchors and limits the influence of returns outside the grid.

For each representative, we evaluate multiscale Gaussian responses between $\mathbf{r}_{t,m}$ and every anchor.
The response at scale $\ell$ is
\begin{equation}
    \kappa_{t,m,s}^{(\ell)}
    =
    \exp\!\left(
        -\tfrac{1}{2}
        (\mathbf{r}_{t,m}-\mathbf{a}_s)^\top
        \Sigma_{\ell}^{-1}
        (\mathbf{r}_{t,m}-\mathbf{a}_s)
    \right),
    \label{eq:gaussian_response}
\end{equation}
The diagonal matrix $\Sigma_{\ell}$ controls spatial coverage along each axis, and $R$ is the number of scales.
We combine and normalize responses across scales to obtain soft assignment weights $\alpha_{t,m,s}$.
These weights collect local returns and wider context within each region.

Each region pools representative features with these soft assignments.
We concatenate the pooled feature with observation support $w_{t,s}=\sum_m\alpha_{t,m,s}$
and the weighted center $\bar{\mathbf{r}}_{t,s}=\sum_m\alpha_{t,m,s}\mathbf{r}_{t,m}/w_{t,s}$.
A shared FFN then produces the slot feature:
\begin{equation}
    \mathbf{z}_{t,s}
    =
    \mathrm{FFN}
    \!\left(
        \sum_{m}\alpha_{t,m,s}\,\mathbf{h}_{t,m}
        \oplus
        \bigl[w_{t,s},\;\bar{\mathbf{r}}_{t,s}\bigr]
    \right).
    \label{eq:slot_readout}
\end{equation}
If $w_{t,s}=0$, no return supports the region, and a learnable bias determines $\mathbf{z}_{t,s}$.
Each observation group yields a fixed-size representation
$Z_t=[\mathbf{z}_{t,1},\ldots,\mathbf{z}_{t,S}]^\top\in\mathbb{R}^{S\times d}$.
Fixed region indices make slot features comparable across time, while their contents capture local state changes.

\subsection{Compositional Motion Modeling}
\label{sec:compositional_motion}

Following the hypothesis in the introduction, we represent local human states as combinations of reusable latent elements.
We introduce a shared prototype bank
$\mathbf{C}=[\mathbf{c}_1,\ldots,\mathbf{c}_K]^\top\in\mathbb{R}^{K\times d}$.
Each prototype is a learnable latent state primitive.
All spatial slots use the same bank.
Each region selects prototypes and weights them to represent its current state.
This avoids assigning a local state to a single prototype.
The resulting compositions can be reused across regions and time without a point cloud reconstruction decoder.

For each slot feature $\mathbf{z}_{t,s}$, we compute its cosine similarity to the prototypes in $\mathbf{C}$, 
and retain the top $k$ prototypes to form a sparse distribution.
Their weighted mixture produces a composition token:
\begin{equation}
    \begin{aligned}
        \mathbf{q}_{t,s}
        &=
        \operatorname{Norm}_1\!\left(
            \operatorname{Top}_{k}\!\left[
                \operatorname{softmax}\!\left(
                    \frac{\operatorname{sim}\{\mathbf{z}_{t,s},\,\operatorname{sg}(\mathbf{C})\}}{\tau}
                \right)
            \right]
        \right), \\
        \mathbf{u}_{t,s}
        &=
        \sqrt{d}\;
        \operatorname{Norm}_2\!\left(\operatorname{sg}(\mathbf{C})^\top\mathbf{q}_{t,s}\right).
    \end{aligned}
    \label{eq:prototype_composition}
\end{equation}
Here, $\tau$ is the assignment temperature.
The operators $\operatorname{Norm}_1$ and $\operatorname{Norm}_2$ denote $\ell_1$ and $\ell_2$ normalization, respectively.
The operator $\operatorname{sg}(\cdot)$ stops gradients, and $\mathbf{C}\in\mathbb{R}^{K\times d}$ is the shared prototype bank.
Each prototype is $\ell_2$-normalized before use.
Thus, $\mathbf{q}_{t,s}$ contains sparse composition weights for the local state.
The corresponding token $\mathbf{u}_{t,s}$ serves as input to the temporal model.
The tokens in one group form $U_t\in\mathbb{R}^{S\times d}$ and retain separate representations for all regions.

We model the temporal evolution of regional compositions with a group-level causal Transformer $\mathcal{A}$.
The input is shifted by one observation group.
Fixed slot position embeddings identify spatial regions.
Group-level rotary position embeddings encode temporal order.
Slots within a group can attend to one another.
A causal mask restricts attention across groups.
Thus, a prediction at position $t$ cannot access observations at time $t$ or later.
To process sequences of arbitrary length, $\mathcal{A}$ maintains a fixed-capacity history cache.
At each window boundary, we stop gradients through the cache and pass it to the next window.

The EMA encoder $E_{\bar\theta}$ provides targets for future groups.
Its parameters track the online encoder through $\bar\theta\leftarrow m\bar\theta+(1-m)\theta$.
For the next group $X_{t+1}$, the EMA encoder produces $\bar{Z}_{t+1}$.
We apply the same sparse assignment from Eq.~(\ref{eq:prototype_composition}) to obtain the stop-gradient target $\bar{\mathbf{q}}_{t+1,s}$.
The AR model matches its prediction $\hat{\mathbf{q}}_{t+1,s}$ to this target for each slot.
This links historical compositions to future observations in the shared prototype space.

\subsection{Pre-Training Objectives}
\label{sec:training_objectives}

The pretraining objective has three terms:
\begin{equation}
    \mathcal{L}
    =
    \mathcal{L}_{\mathrm{AR}}
    + \lambda_1 \mathcal{L}_{\mathrm{pro}}
    + \lambda_2 \mathcal{L}_{\mathrm{con}},
    \label{eq:loss_total}
\end{equation}
$\mathcal{L}_{\mathrm{AR}}$ and $\mathcal{L}_{\mathrm{con}}$ update the online encoder.
$\mathcal{L}_{\mathrm{pro}}$ updates only the prototype bank $\mathbf{C}$ through a separate optimizer, and the two paths share no gradients.
Here, $\lambda_1$ and $\lambda_2$ weight the corresponding terms.
We describe each term below.

\textbf{Prototype temporal prediction loss.}
We use historical compositions to predict the prototype distribution of the next observation group.
This objective captures temporal evolution and coordination across regions.
The model takes composition tokens from a fixed window and a cached history as input.
A cosine prediction head followed by a temperature-scaled softmax maps the temporal features to $\hat{\mathbf{q}}_{t+1,s}$.
The EMA features of the next group query the shared prototype bank $\mathbf{C}$.
The sparse assignment in Eq.~(\ref{eq:prototype_composition}) then gives the target $\bar{\mathbf{q}}_{t+1,s}$.
We use per-slot soft-target cross-entropy to learn future element selection and mixture weights:
\begin{equation}
    \mathcal{L}_{\mathrm{AR}}
    =
    -\mathbb{E}
    \left[
        \sum_{j=1}^{K}
        \bar{q}_{t+1,s,j}
        \log \hat{q}_{t+1,s,j}
    \right].
    \label{eq:loss_ar}
\end{equation}
Here, $\mathbb{E}[\cdot]$ averages over valid samples.
These include all prediction positions with a next group ($t<T$) and all spatial slots $s$.
Gradients do not pass through the target $\bar{\mathbf{q}}_{t+1,s}$.
This loss updates the online encoder and temporal prediction module.
The prototype bank $\mathbf{C}$ receives no gradient.

\textbf{Prototype bank learning loss.}
We learn the shared prototype bank $\mathbf{C}$ through a separate balanced assignment objective.
This encourages reusable latent elements and limits concentration on a few prototypes.
We compute cosine similarities between sampled EMA slot features $\bar{\mathbf{z}}_{t,s}$ and the prototypes in $\mathbf{C}$.
An untruncated softmax gives the distribution $\boldsymbol{\rho}_{t,s}$.
For each spatial slot, we also apply Sinkhorn balancing to these similarities across sampled features.
This produces the target distribution $\mathbf{b}_{t,s}$ and encourages more balanced prototype use.
We match the ordinary assignments to these targets with cross-entropy:
\begin{equation}
    \mathcal{L}_{\mathrm{pro}}
    =
    -\mathbb{E}
    \left[
        \sum_{j=1}^{K}
        b_{t,s,j}\log \rho_{t,s,j}
    \right].
    \label{eq:loss_pro}
\end{equation}
Here, $\mathbb{E}[\cdot]$ averages over sampled time steps $t$ and spatial slots $s$ used for prototype learning.
The terms $b_{t,s,j}$ and $\rho_{t,s,j}$ are the $j$th entries of $\mathbf{b}_{t,s}$ and $\boldsymbol{\rho}_{t,s}$, respectively.
Gradients do not pass through the EMA features or Sinkhorn targets.
This loss therefore updates only the prototype bank $\mathbf{C}$.
Balanced assignment organizes the prototype space.
Temporal prediction still uses the sparse composition distributions as targets.

\textbf{Observation consistency loss.}
We perturb each observation group through local point dropout or the addition of spurious echoes.
The added echoes are shifted radially.
We align the perturbed and clean spatial representations to reduce sensitivity to missing returns and spurious echoes.
Both views share the same spatial slot layout, so we can match corresponding regions directly:
\begin{equation}
    \widetilde{Z}_t = E_{\theta}(\Phi(X_t)),
    \qquad
    \bar{Z}_t = \operatorname{sg}
    \left(E_{\bar{\theta}}(X_t)\right).
    \label{eq:consistency_views}
\end{equation}
Here, $E_{\theta}$ is the online encoder, and $E_{\bar{\theta}}$ is the EMA encoder.
The term $\Phi(X_t)$ denotes the perturbed observation, and $\operatorname{sg}(\cdot)$ stops gradients.

Regions differ in the amount of valid radar support.
We therefore weight them by the support $w'_{t,s}$ from the clean observation.
The resulting loss is a weighted SmoothL1 loss:
\begin{equation}
    \mathcal{L}_{\mathrm{con}}
    =
    \mathbb{E}
    \left[
        \frac{
            \sum_{s\in\mathcal{S}_t}
            w'_{t,s}\,
            \ell_s
            \left(
                \widetilde{\mathbf z}_{t,s},
                \bar{\mathbf z}_{t,s}
            \right)
        }{
            \sum_{s\in\mathcal{S}_t}w'_{t,s}
        }
    \right],
    \label{eq:loss_con}
\end{equation}
The function $\ell_s$ denotes SmoothL1 after standardizing both representations with fixed per-slot, per-feature statistics.
Here, $\mathbb{E}[\cdot]$ averages over observation groups used for consistency training.
The set $\mathcal{S}_t=\{s\mid w'_{t,s}>0\}$ contains regions with radar support in the current group.
We obtain $w'_{t,s}$ by applying a saturating transform to each scale's support and averaging across spatial scales.
The weight $w'_{t,s}$ measures support from the clean observation only.
This prevents dropout in the perturbed view from weakening supervision of the same region.
The features $\widetilde{\mathbf{z}}_{t,s}$ and $\bar{\mathbf{z}}_{t,s}$ come from the perturbed and clean views, respectively.
This objective acts directly on continuous spatial representations.
It encourages stable local states despite random missing detections or spurious echoes.

\section{Experiment}
\subsection{Datasets}

We survey public datasets for mmWave human sensing, including M4Human~\citep{fan2026m4human},
DGUHA~\citep{lee2023dguha}, MARS~\citep{an2021mars}, milliFlow~\citep{ding2024milliflow},
MiliPoint~\citep{cui2023milipoint}, and mmBody~\citep{chen2022mmbody}.
The table also lists mRI~\citep{an2022mri}, MM-Fi~\citep{yang2023mmfi}, and RadHAR~\citep{singh2019radhar}.
These three datasets are the sources of UniMM-HAR~\citep{lin2026dap}.
Table~\ref{tab:datasets} summarizes their modalities, annotations, and sizes.
Appendix~\ref{app:dataset} provides further details.

\begin{table}[t]
\centering
\caption{Statistics of the mmWave human datasets used in our experiments.
RF-BW reports the radar chirp operation band.
The upper block lists pretraining data; the lower block covers downstream evaluation.
$^{\dagger}$UniMM-HAR unifies mRI, MM-Fi, and RadHAR for action classification.}
\label{tab:datasets}
\small
\setlength{\tabcolsep}{3pt}
\resizebox{\textwidth}{!}{%
\begin{tabular}{l c | ccc | ccc | rrrr}
\toprule
\multicolumn{2}{c|}{\textbf{Dataset}} &
\multicolumn{3}{c|}{\textbf{Modalities}} &
\multicolumn{3}{c|}{\textbf{Annotations}} &
\multicolumn{4}{c}{\textbf{Statistics}} \\
\cmidrule(r){1-2}
\cmidrule(lr){3-5}
\cmidrule(lr){6-8}
\cmidrule(l){9-12}
Name & RF-BW (GHz) &
RGB & Depth & mmWave &
Actions & Skeleton & Mesh &
\# Act. & \# Sub. & \# Seq. & \# Frame \\
\midrule
M4Human & 62--69
& $\checkmark$ & $\checkmark$ & $\checkmark$
& $\checkmark$ & $\checkmark$ & $\checkmark$
& 50 & 20 & 999 & 661K \\
\addlinespace[1.5pt]
MARS & 77--80.2
& -- & -- & $\checkmark$
& $\checkmark$ & $\checkmark$ & --
& 10 & 4 & 80 & 40K \\
\addlinespace[1.5pt]
milliFlow & 62--63.6
& $\checkmark$ & $\checkmark$ & $\checkmark$
& $\checkmark$ & $\checkmark$ & --
& 8 & 12 & -- & 43.2K \\
\addlinespace[1.5pt]
MiliPoint & 77--81
& -- & -- & $\checkmark$
& $\checkmark$ & $\checkmark$ & --
& 49 & 11 & -- & 545K \\
\midrule
DGUHA & 76--81
& -- & -- & $\checkmark$
& $\checkmark$ & $\checkmark$ & --
& 7 & 19 & -- & -- \\
\addlinespace[1.5pt]
mmBody & 76--81
& $\checkmark$ & $\checkmark$ & $\checkmark$
& $\checkmark$ & $\checkmark$ & $\checkmark$
& 100 & 20 & -- & $>$200K \\
\addlinespace[1.5pt]
mRI & 77--80.2
& $\checkmark$ & $\checkmark$ & $\checkmark$
& $\checkmark$ & $\checkmark$ & --
& 12 & 20 & 300 & 160K \\
\addlinespace[1.5pt]
MM-Fi & 60--64
& $\checkmark$ & $\checkmark$ & $\checkmark$
& $\checkmark$ & $\checkmark$ & --
& 27 & 40 & 1080 & $>$320K \\
\addlinespace[1.5pt]
RadHAR & 76--81
& -- & -- & $\checkmark$
& $\checkmark$ & -- & --
& 5 & 2 & -- & $\sim$167K \\
\addlinespace[1.5pt]
UniMM-HAR$^{\dagger}$ & 60--64 / 76--81
& -- & -- & $\checkmark$
& $\checkmark$ & -- & --
& 33 & 62 & 40,494 & $\sim$1.30M \\
\bottomrule
\end{tabular}}
\end{table}

\textbf{Tasks and Data Splits.}
Pretraining uses unlabeled mmWave point cloud sequences from M4Human, milliFlow, MARS, and MiliPoint.
None of the downstream evaluation data enter pretraining.
We select four representative downstream tasks, \textbf{(1) action classification, (2) dense point generation, (3) joint pose estimation}, and \textbf{(4) future motion prediction}.
Action classification assigns an action label to each radar point cloud sequence under the official C-Sub (cross-subject) and C-Set (cross-scene) protocols of UniMM-HAR.
Dense point generation produces 512 3D points from a sparse radar frame, following the mmWaveFlow~\citep{su2026mmwaveflow} split on mmBody.
The test set includes rain and low-light scenes.
We exclude smoke and occlusion scenes because their dense point clouds are severely degraded or missing.
Joint pose estimation regresses current 3D joint positions under the official Random, Cross-Subject, and Cross-Environment protocols of MM-Fi.
Future motion prediction follows the cross-subject split of mmPred~\citep{fan2026mmpred} on MM-Fi.
It predicts ten frames of joint trajectories from eight radar frames.

\subsection{Experiment Settings}
\label{sec:exp_settings}

(1) \textbf{Pre-Training.}
Different radar devices produce different frame rates and numbers of points per frame.
We resample frame sequences to 12Hz.
We express point coordinates in meters with the radar as the origin.
We normalize Doppler values to $[-1,1]$ using per-frame min--max statistics.
We normalize reflection intensity in the logarithmic (dB) domain and mark missing attributes with separate masks.
The four sources contain approximately 1.29M raw frames.
Pretraining is fully self-supervised.
It uses only point cloud sequences, without action, skeleton, or mesh labels.

We pretrain for 100 epochs with batch size 32 on two NVIDIA H800 GPUs using BF16 mixed precision and \texttt{torch.compile}.
M4Human is resampled each epoch to balance sources.
The online encoder and AR module use AdamW with learning rate $10^{-4}$ and weight decay 0.05.
A separate optimizer trains the prototype bank with the same learning rate and weight decay 0.01.
Both optimizers clip gradient norms at 1.0.
We use five epochs of linear warmup followed by cosine decay to $10^{-5}$.
The initial EMA momentum is 0.996, and AR memory holds 32 observation groups.

(2) \textbf{Downstream.}
All four downstream tasks share the same frozen pretrained encoder; we fit only lightweight task-specific decoders.
We use a single-layer linear probe for action classification, a ResNet decoder for dense point generation, and four-block TCN decoders for joint pose estimation and future motion prediction.
Implementation details are provided in Appendices~\ref{app:har}--\ref{app:motion}.
Appendix~\ref{app:metrics} defines the evaluation metrics and reporting conventions.

\textbf{Baselines.}
Our general point cloud baselines include
Point-MAE~\citep{pang2022pointmae}, Point-BERT~\citep{yu2022pointbert},
Point-M2AE~\citep{zhang2022pointm2ae}, and PointGPT-S~\citep{chen2023pointgpt}.
We also evaluate Sonata~\citep{wu2025sonata} using its publicly released pretrained encoder.
The self-supervised radar tokenizer of Wave2Body~\citep{liang2026wave2body} (W2B in the tables) serves as a radar-specific baseline.
Within each downstream task, all pretrained encoders use the same data split and prediction head.
Appendix~\ref{app:baselines} describes their designs and implementation details.

\subsection{Experiment Results}

Tables~\ref{tab:har}--\ref{tab:pred} summarize results on the four downstream tasks.
PT$^3$ reaches 81.35\% and 85.23\% accuracy on UniMM-HAR under C-Sub and C-Set.
On mmBody, it obtains CD/EMD of 0.91/4.70.
On MM-Fi, it obtains 77.43\,mm MPJPE under Cross-Environment.
\begin{table}[t]
\centering
\begin{minipage}{\textwidth}
\centering
\captionof{table}{Action classification accuracy (\%) on UniMM-HAR.}
\label{tab:har}
\scriptsize
\setlength{\tabcolsep}{2pt}
\resizebox{\textwidth}{!}{%
\begin{tabular}{l *{3}{c} @{\hspace{2pt}\vrule width 0.4pt\hspace{2pt}} *{6}{c} @{\hspace{5pt}} c}
\toprule
\textbf{Method} & RadHAR & FastHAR & DAPNet & PointMAE & PointM2AE & PointBERT & PointGPT & Sonata & W2B & \cellcolor{blue!8}\textbf{PT$^3$} \\
\midrule
C-Sub$\uparrow$ & 42.49 & 53.72 & 80.72 & 61.68 & 65.65 & 76.89 & 56.48 & 72.44 & 78.49 & \cellcolor{blue!8}\textbf{81.35} \\
C-Set$\uparrow$ & 48.47 & 61.16 & 81.82 & 60.11 & 59.48 & 76.17 & 53.19 & 75.34 & 69.93 & \cellcolor{blue!8}\textbf{85.23} \\
\bottomrule
\end{tabular}}
\end{minipage}

\vspace{8pt}

\begin{minipage}{\textwidth}
\centering
\captionof{table}{Dense point generation on mmBody.}
\label{tab:dense}
\scriptsize
\setlength{\tabcolsep}{1pt}
\newcommand{\denseval}[1]{{\fontsize{8.5pt}{10pt}\selectfont #1}}
\resizebox{\textwidth}{!}{%
\begin{tabular}{l *{6}{c} @{\hspace{2pt}\vrule width 0.4pt\hspace{2pt}} *{6}{c} @{\hspace{5pt}} c}
\toprule
\textbf{Method} & mmPoint & RadarHD & RadarDiff & LiDiff & Tiger & mmWaveFlow & PointMAE & PointM2AE & PointBERT & PointGPT & Sonata & W2B & \cellcolor{blue!8}\textbf{PT$^3$} \\
\midrule
CD$\downarrow$ & \denseval{1.09} & \denseval{1.59} & \denseval{1.09} & \denseval{1.16} & \denseval{1.02} & \denseval{0.95} & \denseval{0.99} & \denseval{0.98} & \denseval{1.00} & \denseval{1.01} & \denseval{1.05} & \denseval{0.97} & \cellcolor{blue!8}\denseval{\textbf{0.91}} \\
EMD$\downarrow$ & \denseval{9.28} & \denseval{14.89} & \denseval{7.11} & \denseval{12.27} & \denseval{5.82} & \denseval{4.76} & \denseval{5.00} & \denseval{5.01} & \denseval{5.00} & \denseval{5.06} & \denseval{4.87} & \denseval{4.91} & \cellcolor{blue!8}\denseval{\textbf{4.70}} \\
\bottomrule
\end{tabular}}
\end{minipage}
\end{table}

\begin{table}[t]
\centering
\caption{Joint pose estimation on MM-Fi. We report MPJPE and PA-MPJPE (PA) in mm, under the Random, Cross-Subject and Cross-Environment protocols.}
\label{tab:pose}
\small
\setlength{\tabcolsep}{4pt}
\renewcommand{\arraystretch}{1.0}
\begin{tabular}{l c c c c c c}
\toprule
& \multicolumn{2}{c}{\textbf{Random}} &
\multicolumn{2}{c}{\textbf{Cross-Subject}} &
\multicolumn{2}{c}{\textbf{Cross-Environment}} \\
\cmidrule(lr){2-3} \cmidrule(lr){4-5} \cmidrule(lr){6-7}
\textbf{Method} &
MPJPE$\downarrow$ & PA$\downarrow$ & MPJPE$\downarrow$ & PA$\downarrow$ & MPJPE$\downarrow$ & PA$\downarrow$ \\
\midrule
PointTransformer & 73.09 & 55.60 & 75.96 & 58.70 & 88.28 & 68.79 \\
DiffPose & 73.44 & 56.83 & 70.31 & 54.12 & 86.35 & 66.87 \\
mmDiff & 68.62 & 53.11 & 68.46 & 52.55
& 85.63 & 66.43 \\
\midrule
PointMAE & 70.91 & 52.71 & 70.15
& 53.77 & 87.48 & 67.69 \\
PointBERT & 70.80 & 52.90 & 70.10
& 53.06 & 83.32 & 63.46 \\
PointM2AE & 72.93 & 55.19 & 72.19
& 55.07 & 89.65 & 69.63 \\
PointGPT & 73.37 & 55.23 & 73.84
& 56.12 & 90.72 & 69.25 \\
Sonata & 75.60 & 58.94 & 72.65 & 57.49 & 85.63 & 67.34 \\
W2B & 68.10 & 51.09 & 67.03
& 51.10 & 82.10 & 63.08 \\
\midrule
\rowcolor{blue!8}
PT$^3$ (ours)
& \textbf{66.74} & \textbf{50.95} & \textbf{65.75} & \textbf{50.92} & \textbf{77.43} & \textbf{58.93} \\
\bottomrule
\end{tabular}
\end{table}

\begin{table}[t]
\centering
\caption{Future motion prediction on MM-Fi for different actions,
evaluated by ADE and FDE.}
\label{tab:pred}
\small
\setlength{\tabcolsep}{3pt}
\renewcommand{\arraystretch}{0.9}
\resizebox{\textwidth}{!}{%
\begin{tabular}{l *{14}{c}}
\toprule
& \multicolumn{2}{c}{Chest Expand} & \multicolumn{2}{c}{Side Twist} & \multicolumn{2}{c}{Raise Hand} & \multicolumn{2}{c}{Pickup} & \multicolumn{2}{c}{Throwing} & \multicolumn{2}{c}{Kicking} & \multicolumn{2}{c}{Bowing} \\
\cmidrule(lr){2-3} \cmidrule(lr){4-5} \cmidrule(lr){6-7} \cmidrule(lr){8-9} \cmidrule(lr){10-11} \cmidrule(lr){12-13} \cmidrule(lr){14-15}
\textbf{Method} & ADE$\downarrow$ & FDE$\downarrow$ & ADE$\downarrow$ & FDE$\downarrow$ & ADE$\downarrow$ & FDE$\downarrow$ & ADE$\downarrow$ & FDE$\downarrow$ & ADE$\downarrow$ & FDE$\downarrow$ & ADE$\downarrow$ & FDE$\downarrow$ & ADE$\downarrow$ & FDE$\downarrow$ \\
\midrule
SPGSN & 0.402 & 0.422 & 0.430 & 0.445 & 0.397 & 0.446 & 0.595 & 0.686 & 0.449 & 0.517 & 0.426 & 0.449 & 0.336 & 0.395 \\
HumanMAC & 0.353 & 0.306 & 0.407 & 0.392 & 0.363 & 0.333 & 0.547 & 0.578 & 0.439 & 0.471 & 0.425 & 0.421 & 0.304 & 0.314 \\
mmPred & \textbf{0.272} & \textbf{0.230} & \textbf{0.315} & \textbf{0.308} & \textbf{0.237} & \textbf{0.218} & 0.452 & 0.481 & 0.371 & 0.370 & 0.374 & 0.368 & 0.246 & 0.234 \\
\midrule
PointMAE & 0.328 & 0.279 & 0.431 & 0.393 & 0.290 & 0.251 & 0.460 & 0.448 & 0.413 & 0.410 & 0.355 & 0.350 & 0.248 & 0.246 \\
PointBERT & 0.310 & 0.263 & 0.376 & 0.343 & 0.275 & 0.246 & 0.449 & 0.447 & 0.394 & 0.399 & 0.356 & 0.354 & 0.240 & 0.242 \\
PointM2AE & 0.370 & 0.325 & 0.441 & 0.405 & 0.322 & 0.277 & 0.459 & 0.453 & 0.437 & 0.422 & 0.376 & 0.365 & 0.249 & 0.245 \\
PointGPT & 0.348 & 0.307 & 0.444 & 0.396 & 0.306 & 0.263 & 0.499 & 0.476 & 0.436 & 0.425 & 0.374 & 0.363 & 0.274 & 0.263 \\
Sonata & 0.344 & 0.275 & 0.452 & 0.412 & 0.398 & 0.351 & 0.547 & 0.534 & 0.453 & 0.443 & 0.418 & 0.405 & 0.298 & 0.289 \\
W2B & 0.311 & 0.258 & 0.363 & 0.334 & 0.268 & 0.245 & 0.441 & 0.432 & 0.393 & 0.386 & 0.349 & 0.344 & 0.226 & 0.226 \\
\midrule
\rowcolor{blue!8}
PT$^3$ (ours) & 0.280 & 0.236 & 0.335 & 0.313 & 0.247 & 0.221 & \textbf{0.408} & \textbf{0.418} & \textbf{0.364} & \textbf{0.357} & \textbf{0.335} & \textbf{0.329} & \textbf{0.223} & \textbf{0.223} \\
\bottomrule
\end{tabular}}
\end{table}

\textbf{Results Discussion.}
PT$^3$ outperforms every compared pretrained baseline across the four tasks.
For cross-scene action classification, its C-Set accuracy exceeds the strongest pretrained baseline by 9.06 percentage points.
For cross-environment joint pose estimation, it reduces MPJPE from 82.10 to 77.43\,mm.
For dense point generation, CD/EMD fall from 0.97/4.91 to 0.91/4.70.

PT$^3$ also surpasses the task-specific supervised methods on all reported metrics for action classification, dense point generation, and joint pose estimation.
For future motion prediction, PT$^3$ achieves lower ADE and FDE than mmPred on Pickup, Throwing, Kicking, and Bowing.
mmPred remains stronger on Chest Expand, Side Twist, and Raise Hand.
Together, these results demonstrate the transferability of the pretrained representations across diverse human sensing tasks.

\section{Ablation and Analysis}

We construct three ablation variants using the same pre-training data and downstream protocols.
Removing observation consistency or autoregressive prediction changes only the corresponding loss term.
Removing prototype modeling also removes the prototype bank and its sole update signal, $\mathcal{L}_{\mathrm{pro}}$.
This variant predicts continuous slot features directly.
We pre-train all variants from scratch.
Table~\ref{tab:ablation} reports the results on four downstream tasks.

\begin{table}[t]
\centering
\caption{Pre-training component ablation across four downstream tasks.}
\label{tab:ablation}
\scriptsize
\setlength{\tabcolsep}{4pt}
\resizebox{\textwidth}{!}{%
\begin{tabular}{l cc cc ccc cc}
\toprule
& \multicolumn{2}{c}{\shortstack{\textbf{HAR}}} & \multicolumn{2}{c}{\shortstack{\textbf{Dense Point Gen.}}} &
\multicolumn{3}{c}{\shortstack{\textbf{Joint pose estimation}(MPJPE$\downarrow$\,/\,PA$\downarrow$)}} & \multicolumn{2}{c}{\shortstack{\textbf{Motion Pre.}}} \\
\cmidrule(lr){2-3} \cmidrule(lr){4-5} \cmidrule(lr){6-8} \cmidrule(lr){9-10}
\textbf{Variant} & C-Sub$\uparrow$ & C-Set$\uparrow$ & CD$\downarrow$ & EMD$\downarrow$ & Random & C-Sub & C-Env & ADE$\downarrow$ & FDE$\downarrow$ \\
\midrule
\rowcolor{blue!8}
Full model & 81.35 & 85.23 & 0.91 & 4.70 & 66.74\,/\,50.95 & 65.75\,/\,50.92 & 77.43\,/\,58.93 & 0.313 & 0.299 \\
\midrule
w/o $L_{\mathrm{con}}$ & 82.28 & 81.79 & 0.95 & 5.15 & 91.04\,/\,73.56 & 92.30\,/\,74.70 & 100.62\,/\,80.29 & 0.379 & 0.364 \\
w/o Prototype (Slot-AR) & 78.70 & 80.66 & 0.91 & 5.10 & 89.26\,/\,71.35 & 90.89\,/\,72.90 & 98.81\,/\,78.49 & 0.367 & 0.349 \\
w/o $L_{\mathrm{AR}}$ & 78.53 & 80.28 & 0.95 & 4.83 & 82.03\,/\,64.87 & 83.74\,/\,65.99 & 92.59\,/\,72.04 & 0.370 & 0.353 \\
\bottomrule
\end{tabular}}
\end{table}

\textbf{Ablation on Observation Consistency.} 
In this ablation, C-Sub accuracy rises slightly from 81.35\% to 82.28\%.
C-Set accuracy instead falls from 85.23\% to 81.79\%.
For cross-environment joint pose estimation, MPJPE rises from 77.43\,mm to 100.62\,mm.
For future motion prediction, ADE rises from 0.313\,m to 0.379\,m.
One possible explanation is that consistency brings corresponding spatial slots closer across clean and perturbed views.
This reduces the influence of missed detections and anomalous reflections.
Such stability may matter more for continuous state estimation.

\textbf{Ablation on Prototype Modeling.} Without prototype modeling, C-Set accuracy falls to 80.66\%.
For cross-environment joint pose estimation, MPJPE rises to 98.81\,mm.
For dense point generation, EMD rises from 4.70 to 5.10, while CD remains at 0.91.
Shared prototypes may provide reusable state elements across regions and time.
The temporal model can then predict compositions of these elements.
Direct prediction of continuous slot features may instead track transient radar reflections.

\textbf{Ablation on Temporal Prediction.} Removing the autoregressive loss worsens all four downstream tasks.
C-Set accuracy falls to 80.28\%, and MPJPE for cross-environment joint pose estimation rises to 92.59\,mm.
For future motion prediction, ADE/FDE rise from 0.313/0.299\,m to 0.370/0.353\,m.
Predicting the next observation group may encourage the encoder to retain the current body state and its temporal evolution.
These features can also support action classification, dense point generation, and joint pose estimation.

\textbf{Prototype Composition.}
Shared prototypes should support reusable, action-dependent compositions across subjects.
We test this by comparing Jensen--Shannon (JS) divergence between prototype-assignment compositions for different actions and for the same action across subjects.
Action-pair distances exceed same-action subject-pair distances, 
indicating action-dependent prototype usage that remains consistent across subjects (Appendix~\ref{app:proto_heatmap} and \ref{app:proto_com}).

\begin{figure}[h]
\centering
\begin{minipage}[t]{0.2\textwidth}
\centering
\includegraphics[width=\linewidth]{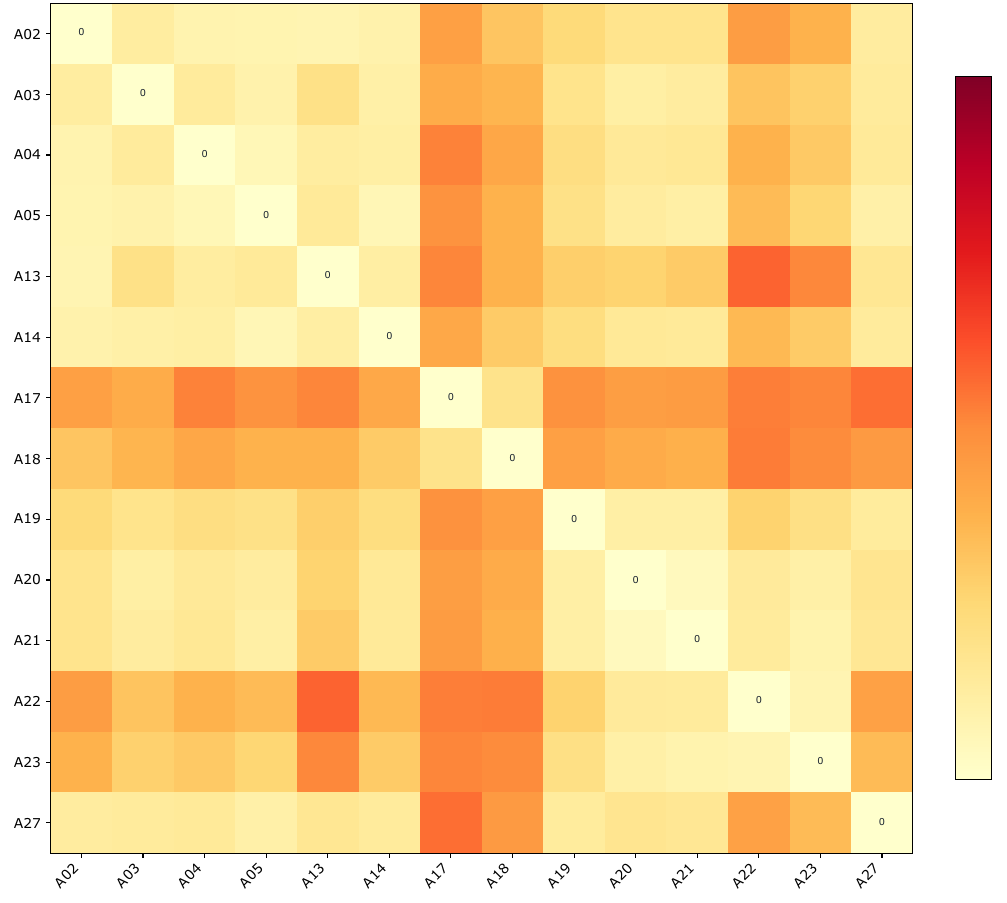}
\end{minipage}
\begin{minipage}[t]{0.2\textwidth}
\centering
\includegraphics[width=\linewidth]{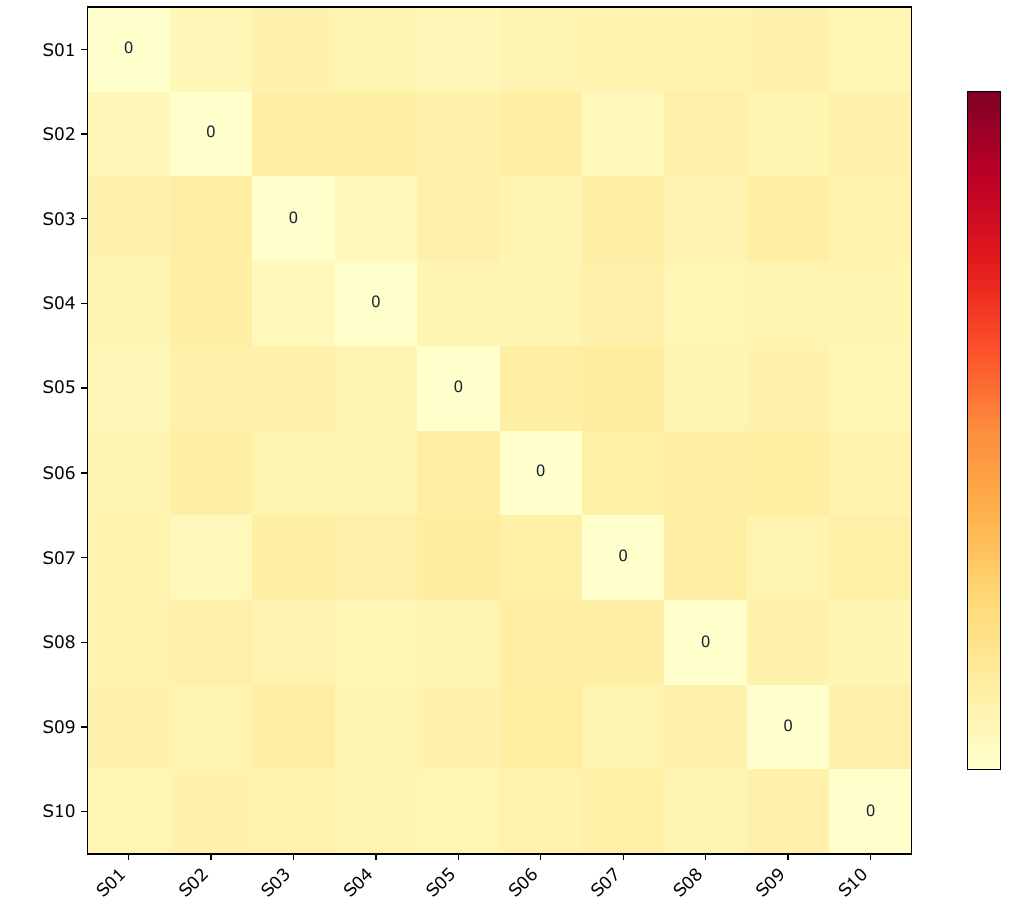}
\end{minipage}
\caption{JS divergence between prototype compositions.
Left: between actions, averaged over subjects.
Right: between subjects, averaged over actions.
Darker cells larger differences.}
\label{fig:proto_js}
\end{figure}

\textbf{Hyper-param sensitivity.} We also analyze sensitivity to the number of spatial slots $S$, top-$k$ prototype selection, and prototype bank size $K$ in Appendix~\ref{app:hyperparameter_analysis}.

\section{Conclusion}
\label{sec:conclusion}

In this paper, we present PT$^3$, a self-supervised pretraining framework for multi-source mmWave human point clouds.
It organizes unstable radar detections into spatial slots and represents local states through shared prototype compositions.
Autoregressive prediction captures how these compositions evolve over time.
Observation consistency encourages stable representations under radar perturbations.
After pretraining on unlabeled radar sequences, the same frozen encoder outperforms all compared pretrained baselines across four downstream tasks.
It also achieves competitive or superior performance to the listed task-specific supervised methods.
Ablations support the contribution of these three designs to transfer performance.
Prototype-composition analysis shows similar assignment patterns for the same action across subjects.
These results support learning transferable human representations from unstable mmWave observations.
Implementation details, additional analyses, and future work are provided in the appendix.

%

\subsection*{Ethics Statement}
This study uses only publicly available mmWave human sensing datasets, all of which have been processed by their original providers.
No new human subject data were collected in this work.


\bibliography{references}

\begin{thebibliography}{47}
\providecommand{\natexlab}[1]{#1}
\providecommand{\url}[1]{\texttt{#1}}
\expandafter\ifx\csname urlstyle\endcsname\relax
  \providecommand{\doi}[1]{doi: #1}\else
  \providecommand{\doi}{doi: \begingroup \urlstyle{rm}\Url}\fi

\bibitem[Abdelsamad et~al.(2025)Abdelsamad, Ulrich, Glaeser, and
  Valada]{abdelsamad2025nomae}
Mohamed Abdelsamad, Michael Ulrich, Claudius Glaeser, and Abhinav Valada.
\newblock Multi-scale neighborhood occupancy masked autoencoder for
  self-supervised learning in {LiDAR} point clouds.
\newblock In \emph{Proceedings of the IEEE/CVF Conference on Computer Vision
  and Pattern Recognition}, pages 22234--22243, 2025.
\newblock URL
  \url{https://openaccess.thecvf.com/content/CVPR2025/html/Abdelsamad_Multi-Scale_Neighborhood_Occupancy_Masked_Autoencoder_for_Self-Supervised_Learning_in_LiDAR_CVPR_2025_paper.html}.

\bibitem[An and Ogras(2021)]{an2021mars}
Sizhe An and Umit~Y. Ogras.
\newblock {MARS}: {mmWave}-based assistive rehabilitation system for smart
  healthcare.
\newblock \emph{ACM Transactions on Embedded Computing Systems}, 20\penalty0
  (5s):\penalty0 1--22, 2021.
\newblock \doi{10.1145/3477003}.
\newblock URL \url{https://doi.org/10.1145/3477003}.

\bibitem[An and Ogras(2022)]{an2022fuse}
Sizhe An and Umit~Y. Ogras.
\newblock Fast and scalable human pose estimation using {mmWave} point cloud.
\newblock In \emph{Proceedings of the Design Automation Conference}, 2022.
\newblock \doi{10.1145/3489517.3530522}.
\newblock URL \url{https://arxiv.org/abs/2205.00097}.

\bibitem[An et~al.(2022)An, Li, and Ogras]{an2022mri}
Sizhe An, Yin Li, and Umit Ogras.
\newblock {mRI}: Multi-modal {3D} human pose estimation dataset using {mmWave},
  {RGB-D}, and inertial sensors.
\newblock In \emph{Advances in Neural Information Processing Systems}, 2022.
\newblock URL \url{https://arxiv.org/abs/2210.08394}.

\bibitem[Chen et~al.(2022)Chen, Wang, Zhu, Li, Chen, and Ye]{chen2022mmbody}
Anjun Chen, Xiangyu Wang, Shaohao Zhu, Yanxu Li, Jiming Chen, and Qi~Ye.
\newblock {mmBody} benchmark: {3D} body reconstruction dataset and analysis for
  millimeter wave radar.
\newblock In \emph{Proceedings of the 30th ACM International Conference on
  Multimedia}, pages 3501--3510, 2022.
\newblock URL \url{https://arxiv.org/abs/2209.05070}.

\bibitem[Chen et~al.(2023{\natexlab{a}})Chen, Wang, Yang, Yu, Yuan, and
  Yue]{chen2023pointgpt}
Guangyan Chen, Meiling Wang, Yi~Yang, Kai Yu, Li~Yuan, and Yufeng Yue.
\newblock {PointGPT}: Auto-regressively generative pre-training from point
  clouds.
\newblock In \emph{Advances in Neural Information Processing Systems},
  volume~36, pages 29667--29679, 2023{\natexlab{a}}.
\newblock \doi{10.52202/075280-1291}.
\newblock URL \url{https://arxiv.org/abs/2305.11487}.

\bibitem[Chen et~al.(2023{\natexlab{b}})Chen, Zhang, Li, Pang, Xia, and
  Liu]{chen2023humanmac}
Ling-Hao Chen, JiaWei Zhang, Yewen Li, Yiren Pang, Xiaobo Xia, and Tongliang
  Liu.
\newblock {HumanMAC}: Masked motion completion for human motion prediction.
\newblock In \emph{Proceedings of the IEEE/CVF International Conference on
  Computer Vision}, pages 9544--9555, 2023{\natexlab{b}}.
\newblock URL
  \url{https://openaccess.thecvf.com/content/ICCV2023/html/Chen_HumanMAC_Masked_Motion_Completion_for_Human_Motion_Prediction_ICCV_2023_paper.html}.

\bibitem[Chen and Yang(2025)]{chen2025xfi}
Xinyan Chen and Jianfei Yang.
\newblock {X-Fi}: A modality-invariant foundation model for multimodal human
  sensing.
\newblock In \emph{International Conference on Learning Representations}, 2025.
\newblock URL \url{https://arxiv.org/abs/2410.10167}.

\bibitem[Cui et~al.(2023)Cui, Zhong, Wu, Shen, Dahnoun, and
  Zhao]{cui2023milipoint}
Han Cui, Shu Zhong, Jiacheng Wu, Zichao Shen, Naim Dahnoun, and Yiren Zhao.
\newblock {MiliPoint}: A point cloud dataset for {mmWave} radar.
\newblock In \emph{Advances in Neural Information Processing Systems}, 2023.
\newblock URL \url{https://arxiv.org/abs/2309.13425}.

\bibitem[Ding et~al.(2024)Ding, Luo, Zhao, and Lu]{ding2024milliflow}
Fangqiang Ding, Zhen Luo, Peijun Zhao, and Chris~Xiaoxuan Lu.
\newblock {milliFlow}: Scene flow estimation on {mmWave} radar point cloud for
  human motion sensing.
\newblock In \emph{European Conference on Computer Vision}, pages 202--221,
  2024.
\newblock \doi{10.1007/978-3-031-72691-0_12}.
\newblock URL \url{https://arxiv.org/abs/2306.17010}.

\bibitem[Fan et~al.(2024)Fan, Yang, Xu, and Xie]{fan2024diffusion}
Junqiao Fan, Jianfei Yang, Yuecong Xu, and Lihua Xie.
\newblock Diffusion model is a good pose estimator from {3D RF-Vision}.
\newblock In \emph{European Conference on Computer Vision}, pages 1--18, 2024.
\newblock \doi{10.1007/978-3-031-72640-8_1}.
\newblock URL
  \url{https://link.springer.com/chapter/10.1007/978-3-031-72640-8_1}.

\bibitem[Fan et~al.(2026{\natexlab{a}})Fan, Rao, Zhang, Yang, and
  Xie]{fan2026mmpred}
Junqiao Fan, Haocong Rao, Jiarui Zhang, Jianfei Yang, and Lihua Xie.
\newblock {mmPred}: Radar-based human motion prediction in the dark.
\newblock \emph{Proceedings of the AAAI Conference on Artificial Intelligence},
  40\penalty0 (5):\penalty0 3777--3785, 2026{\natexlab{a}}.
\newblock \doi{10.1609/aaai.v40i5.37378}.
\newblock URL \url{https://ojs.aaai.org/index.php/AAAI/article/view/37378}.

\bibitem[Fan et~al.(2026{\natexlab{b}})Fan, Zhou, Yang, Cui, Zhang, Xie, Yang,
  Lu, and Ding]{fan2026m4human}
Junqiao Fan, Yunjiao Zhou, Yizhuo Yang, Xinyuan Cui, Jiarui Zhang, Lihua Xie,
  Jianfei Yang, Chris~Xiaoxuan Lu, and Fangqiang Ding.
\newblock {M4Human}: A large-scale multimodal {mmWave} radar benchmark for
  human mesh reconstruction.
\newblock In \emph{Proceedings of the IEEE/CVF Conference on Computer Vision
  and Pattern Recognition (CVPR)}, pages 42836--42846, June 2026{\natexlab{b}}.
\newblock URL
  \url{https://openaccess.thecvf.com/content/CVPR2026/html/Fan_M4Human_A_Large-Scale_Multimodal_mmWave_Radar_Benchmark_for_Human_Mesh_CVPR_2026_paper.html}.

\bibitem[Gong et~al.(2026)Gong, Liang, Zhu, Gao, Chen, Nakazawa, and
  Xu]{gong2026wiswiss}
Chen Gong, Bo~Liang, Qihao Zhu, Wei Gao, Yin Chen, Jin Nakazawa, and Chenren
  Xu.
\newblock Towards generalizable wireless sensing models via pre-training on
  multi-source datasets.
\newblock In \emph{Proceedings of the 2026 ACM/IEEE International Conference on
  Embedded Artificial Intelligence and Sensing Systems}, pages 488--502, 2026.
\newblock \doi{10.1145/3774906.3802759}.
\newblock URL \url{https://sites.pitt.edu/~weigao/publications/sensys26.pdf}.

\bibitem[Gong et~al.(2023)Gong, Foo, Fan, Ke, Rahmani, and
  Liu]{gong2023diffpose}
Jia Gong, Lin~Geng Foo, Zhipeng Fan, Qiuhong Ke, Hossein Rahmani, and Jun Liu.
\newblock {DiffPose}: Toward more reliable {3D} pose estimation.
\newblock In \emph{Proceedings of the IEEE/CVF Conference on Computer Vision
  and Pattern Recognition}, pages 13041--13051, 2023.

\bibitem[Krispel et~al.(2024)Krispel, Schinagl, Fruhwirth-Reisinger, Possegger,
  and Bischof]{krispel2024maeli}
Georg Krispel, David Schinagl, Christian Fruhwirth-Reisinger, Horst Possegger,
  and Horst Bischof.
\newblock {MAELi}: Masked autoencoder for large-scale {LiDAR} point clouds.
\newblock In \emph{Proceedings of the IEEE/CVF Winter Conference on
  Applications of Computer Vision}, pages 3383--3392, 2024.
\newblock URL \url{https://arxiv.org/abs/2212.07207}.

\bibitem[Lai et~al.(2026)Lai, Yang, Xia, Lin, Sun, Wang, Liu, Wu, and
  Pei]{lai2026radarllm}
Zengyuan Lai, Jiarui Yang, Songpengcheng Xia, Lizhou Lin, Lan Sun, Renwen Wang,
  Jianran Liu, Qi~Wu, and Ling Pei.
\newblock {RadarLLM}: Empowering large language models to understand human
  motion from millimeter-wave point cloud sequence.
\newblock \emph{Proceedings of the AAAI Conference on Artificial Intelligence},
  40\penalty0 (7):\penalty0 5791--5799, 2026.
\newblock \doi{10.1609/aaai.v40i7.37500}.
\newblock URL \url{https://ojs.aaai.org/index.php/AAAI/article/view/37500}.

\bibitem[Lee and Kim(2023)]{lee2023dguha}
Gawon Lee and Jihie Kim.
\newblock {MTGEA}: A multimodal two-stream {GNN} framework for efficient point
  cloud and skeleton data alignment.
\newblock \emph{Sensors}, 23\penalty0 (5):\penalty0 2787, 2023.
\newblock \doi{10.3390/s23052787}.
\newblock URL \url{https://doi.org/10.3390/s23052787}.

\bibitem[Li et~al.(2022)Li, Chen, Zhang, Xie, Tian, and Zhang]{li2022psgsn}
Maosen Li, Siheng Chen, Zijing Zhang, Lingxi Xie, Qi~Tian, and Ya~Zhang.
\newblock Skeleton-parted graph scattering networks for {3D} human motion
  prediction.
\newblock In \emph{European Conference on Computer Vision}, 2022.
\newblock \doi{10.1007/978-3-031-20068-7_2}.
\newblock URL
  \url{https://www.ecva.net/papers/eccv_2022/papers_ECCV/html/4239_ECCV_2022_paper.php}.

\bibitem[Liang et~al.(2026{\natexlab{a}})Liang, Gong, Gao, and
  Xu]{liang2026wave2body}
Bo~Liang, Chen Gong, Wei Gao, and Chenren Xu.
\newblock {Wave2Body}: Rethinking {mmWave} human pose estimation as
  radar-to-body token translation.
\newblock \emph{arXiv preprint arXiv:2607.18875}, 2026{\natexlab{a}}.
\newblock URL \url{https://arxiv.org/abs/2607.18875}.

\bibitem[Liang et~al.(2026{\natexlab{b}})Liang, Gong, Wang, Liu, Zhou, Shao,
  Wang, Gao, Zhou, Cui, and Xu]{liang2026wicompass}
Bo~Liang, Chen Gong, Haobo Wang, Qirui Liu, Rungui Zhou, Fengzhi Shao, Yubo
  Wang, Wei Gao, Kaichen Zhou, Guolong Cui, and Chenren Xu.
\newblock {WiCompass}: Oracle-driven data scaling for {mmWave} human pose
  estimation.
\newblock In \emph{Proceedings of the 32nd Annual International Conference on
  Mobile Computing and Networking}, 2026{\natexlab{b}}.
\newblock \doi{10.1145/3795866.3796684}.
\newblock URL \url{https://arxiv.org/abs/2602.18726}.

\bibitem[Lin et~al.(2026)Lin, Wu, Liu, Wang, and Liu]{lin2026dap}
Jiaying Lin, Shiman Wu, Jinfu Liu, Can Wang, and Mengyuan Liu.
\newblock {DAP}: Doppler-aware point network for heterogeneous {mmWave} action
  recognition.
\newblock In \emph{European Conference on Computer Vision}, 2026.
\newblock URL \url{https://eccv2024.ecva.net/virtual/2026/poster/3612}.

\bibitem[Liu et~al.(2026{\natexlab{a}})Liu, He, Liang, Huang, Zhong, and
  Wu]{liu2026mmjepaecg}
Ziyang Liu, Siyuan He, Feng Liang, Chang Huang, Shuxin Zhong, and Kaishun Wu.
\newblock {mmJEPA-ECG}: Cross-posture robust contactless electrocardiogram
  monitoring via millimeter wave radar sensing.
\newblock \emph{Proceedings of the AAAI Conference on Artificial Intelligence},
  40\penalty0 (45):\penalty0 38962--38970, 2026{\natexlab{a}}.
\newblock \doi{10.1609/aaai.v40i45.41242}.
\newblock URL \url{https://ojs.aaai.org/index.php/AAAI/article/view/41242}.

\bibitem[Liu et~al.(2026{\natexlab{b}})Liu, Xiao, Huang, Yu, He, and
  Wu]{liu2026cardiommflow}
Ziyang Liu, Ruiqiang Xiao, Chang Huang, Kieren Yu, Siyuan He, and Kaishun Wu.
\newblock {Cardio-mmFlow}: A gaussian-prior-free physics-informed flow matching
  framework for electrocardiogram to {mmWave} radar synthesis.
\newblock In \emph{Proceedings of the 43rd International Conference on Machine
  Learning}, 2026{\natexlab{b}}.
\newblock URL \url{https://openreview.net/forum?id=e43JwTIb2V}.

\bibitem[Min et~al.(2024)Min, Xiao, Zhao, Nie, and Dai]{min2024occupancymae}
Chen Min, Liang Xiao, Dawei Zhao, Yiming Nie, and Bin Dai.
\newblock {Occupancy-MAE}: Self-supervised pre-training large-scale {LiDAR}
  point clouds with masked occupancy autoencoders.
\newblock \emph{IEEE Transactions on Intelligent Vehicles}, 9\penalty0
  (7):\penalty0 5150--5162, 2024.
\newblock \doi{10.1109/TIV.2023.3322409}.
\newblock URL \url{https://doi.org/10.1109/TIV.2023.3322409}.

\bibitem[Nisar and Waslander(2025)]{nisar2025psassl}
Barza Nisar and Steven~L. Waslander.
\newblock {PSA-SSL}: Pose and size-aware self-supervised learning on {LiDAR}
  point clouds.
\newblock In \emph{Proceedings of the IEEE/CVF Conference on Computer Vision
  and Pattern Recognition}, pages 6670--6679, 2025.
\newblock URL \url{https://arxiv.org/abs/2503.13914}.

\bibitem[Nunes et~al.(2022)Nunes, Marcuzzi, Chen, Behley, and
  Stachniss]{nunes2022segcontrast}
Lucas Nunes, Rodrigo Marcuzzi, Xieyuanli Chen, Jens Behley, and Cyrill
  Stachniss.
\newblock {SegContrast}: {3D} point cloud feature representation learning
  through self-supervised segment discrimination.
\newblock \emph{IEEE Robotics and Automation Letters}, 7\penalty0 (2):\penalty0
  2116--2123, 2022.
\newblock URL \url{https://www.ipb.uni-bonn.de/pdfs/nunes2022ral-icra.pdf}.

\bibitem[Nunes et~al.(2024)Nunes, Marcuzzi, Mersch, Behley, and
  Stachniss]{nunes2024lidiff}
Lucas Nunes, Rodrigo Marcuzzi, Benedikt Mersch, Jens Behley, and Cyrill
  Stachniss.
\newblock Scaling diffusion models to real-world {3D} {LiDAR} scene completion.
\newblock In \emph{2024 IEEE/CVF Conference on Computer Vision and Pattern
  Recognition (CVPR)}, pages 14770--14780, 2024.
\newblock \doi{10.1109/CVPR52733.2024.01399}.

\bibitem[Palipana et~al.(2021)Palipana, Salami, Leiva, and
  Sigg]{palipana2021pantomime}
Sameera Palipana, Dariush Salami, Luis~A. Leiva, and Stephan Sigg.
\newblock Pantomime: Mid-air gesture recognition with sparse millimeter-wave
  radar point clouds.
\newblock \emph{Proceedings of the ACM on Interactive, Mobile, Wearable and
  Ubiquitous Technologies}, 5\penalty0 (1):\penalty0 27:1--27:27, 2021.
\newblock \doi{10.1145/3448110}.

\bibitem[Pang et~al.(2022)Pang, Wang, Tay, Liu, Tian, and
  Yuan]{pang2022pointmae}
Yatian Pang, Wenxiao Wang, Francis E.~H. Tay, Wei Liu, Yonghong Tian, and
  Li~Yuan.
\newblock Masked autoencoders for point cloud self-supervised learning.
\newblock In \emph{European Conference on Computer Vision}, pages 604--621,
  2022.
\newblock \doi{10.1007/978-3-031-20086-1_35}.
\newblock URL \url{https://arxiv.org/abs/2203.06604}.

\bibitem[Prabhakara et~al.(2023)Prabhakara, Jin, Das, Bhatt, Kumari,
  Soltanaghai, Bilmes, Kumar, and Rowe]{prabhakara2023radarhd}
Akarsh Prabhakara, Tao Jin, Arnav Das, Gantavya Bhatt, Lilly Kumari, Elahe
  Soltanaghai, Jeff Bilmes, Swarun Kumar, and Anthony Rowe.
\newblock High resolution point clouds from {mmWave} radar.
\newblock In \emph{2023 IEEE International Conference on Robotics and
  Automation (ICRA)}, pages 4135--4142, 2023.
\newblock \doi{10.1109/ICRA48891.2023.10161429}.

\bibitem[Ren et~al.(2024)Ren, Kim, Liu, and Liu]{ren2024tiger}
Zhiyuan Ren, Minchul Kim, Feng Liu, and Xiaoming Liu.
\newblock {TIGER}: Time-varying denoising model for {3D} point cloud generation
  with diffusion process.
\newblock In \emph{2024 IEEE/CVF Conference on Computer Vision and Pattern
  Recognition (CVPR)}, pages 9462--9471, 2024.
\newblock \doi{10.1109/CVPR52733.2024.00904}.

\bibitem[Singh et~al.(2019)Singh, Sandha, Garcia, and
  Srivastava]{singh2019radhar}
Akash~Deep Singh, Sandeep~Singh Sandha, Luis Garcia, and Mani Srivastava.
\newblock {RadHAR}: Human activity recognition from point clouds generated
  through a millimeter-wave radar.
\newblock In \emph{Proceedings of the 3rd ACM Workshop on Millimeter-wave
  Networks and Sensing Systems}, pages 51--56, 2019.
\newblock \doi{10.1145/3349624.3356768}.

\bibitem[Su et~al.(2026)Su, Jin, Shi, and Wang]{su2026mmwaveflow}
Chang Su, Beihong Jin, Qiwen Shi, and Zhi Wang.
\newblock {mmWaveFlow}: Unified enhancement and generation of {mmWave} human
  point clouds.
\newblock In \emph{Proceedings of the IEEE/CVF Conference on Computer Vision
  and Pattern Recognition}, pages 31366--31376, 2026.
\newblock URL
  \url{https://openaccess.thecvf.com/content/CVPR2026/papers/Su_mmWaveFlow_Unified_Enhancement_and_Generation_of_mmWave_Human_Point_Clouds_CVPR_2026_paper.pdf}.

\bibitem[Tian et~al.(2023)Tian, Ran, Wang, and Zhao]{tian2023geomae}
Xiaoyu Tian, Haoxi Ran, Yue Wang, and Hang Zhao.
\newblock {GeoMAE}: Masked geometric target prediction for self-supervised
  point cloud pre-training.
\newblock In \emph{Proceedings of the IEEE/CVF Conference on Computer Vision
  and Pattern Recognition}, pages 13570--13580, 2023.
\newblock URL \url{https://arxiv.org/abs/2305.08808}.

\bibitem[Wei et~al.(2026)Wei, Fang, Chetty, Cho, and
  Bianchi-Berthouze]{wei2026maepose}
Xijia Wei, Yuan Fang, Kevin Chetty, Youngjun Cho, and Nadia Bianchi-Berthouze.
\newblock {MAEPose}: Self-supervised spatiotemporal learning for human pose
  estimation on {mmWave} video.
\newblock \emph{arXiv preprint arXiv:2605.00242}, 2026.
\newblock URL \url{https://arxiv.org/abs/2605.00242}.

\bibitem[Wu et~al.(2023)Wu, Wen, Liu, and Zhao]{wu2023msc}
Xiaoyang Wu, Xin Wen, Xihui Liu, and Hengshuang Zhao.
\newblock Masked scene contrast: A scalable framework for unsupervised {3D}
  representation learning.
\newblock In \emph{Proceedings of the IEEE/CVF Conference on Computer Vision
  and Pattern Recognition}, pages 9415--9424, 2023.
\newblock URL
  \url{https://openaccess.thecvf.com/content/CVPR2023/html/Wu_Masked_Scene_Contrast_A_Scalable_Framework_for_Unsupervised_3D_Representation_CVPR_2023_paper.html}.

\bibitem[Wu et~al.(2025)Wu, DeTone, Frost, Shen, Xie, Yang, Engel, Newcombe,
  Zhao, and Straub]{wu2025sonata}
Xiaoyang Wu, Daniel DeTone, Duncan Frost, Tianwei Shen, Chris Xie, Nan Yang,
  Jakob Engel, Richard Newcombe, Hengshuang Zhao, and Julian Straub.
\newblock Sonata: Self-supervised learning of reliable point representations.
\newblock In \emph{Proceedings of the IEEE/CVF Conference on Computer Vision
  and Pattern Recognition (CVPR)}, pages 22193--22204, June 2025.

\bibitem[Xie et~al.(2023)Xie, Deng, Cheng, Zhao, Patel, Trigoni, and
  Markham]{xie2023mmpoint}
Qian Xie, Qianyi Deng, Ta-Ying Cheng, Peijun Zhao, Amir Patel, Niki Trigoni,
  and Andrew Markham.
\newblock {mmPoint}: Dense human point cloud generation from {mmWave}.
\newblock In \emph{British Machine Vision Conference (BMVC)}, 2023.

\bibitem[Xie et~al.(2020)Xie, Gu, Guo, Qi, Guibas, and
  Litany]{xie2020pointcontrast}
Saining Xie, Jiatao Gu, Demi Guo, Charles~R. Qi, Leonidas~J. Guibas, and
  Or~Litany.
\newblock {PointContrast}: Unsupervised pre-training for {3D} point cloud
  understanding.
\newblock In \emph{Computer Vision -- ECCV 2020}, pages 574--591, 2020.
\newblock URL \url{https://arxiv.org/abs/2007.10985}.

\bibitem[Yang et~al.(2023{\natexlab{a}})Yang, He, Liu, Chen, Wu, Lin, He, and
  Ouyang]{yang2023gdmae}
Honghui Yang, Tong He, Jiaheng Liu, Hua Chen, Boxi Wu, Binbin Lin, Xiaofei He,
  and Wanli Ouyang.
\newblock {GD-MAE}: Generative decoder for {MAE} pre-training on {LiDAR} point
  clouds.
\newblock In \emph{Proceedings of the IEEE/CVF Conference on Computer Vision
  and Pattern Recognition}, pages 9403--9414, 2023{\natexlab{a}}.
\newblock URL \url{https://arxiv.org/abs/2212.03010}.

\bibitem[Yang et~al.(2023{\natexlab{b}})Yang, Huang, Zhou, Chen, Xu, Yuan, Zou,
  Lu, and Xie]{yang2023mmfi}
Jianfei Yang, He~Huang, Yunjiao Zhou, Xinyan Chen, Yuecong Xu, Shenghai Yuan,
  Han Zou, Chris~Xiaoxuan Lu, and Lihua Xie.
\newblock {MM-Fi}: Multi-modal non-intrusive {4D} human dataset for versatile
  wireless sensing.
\newblock In \emph{Advances in Neural Information Processing Systems, Datasets
  and Benchmarks Track}, volume~36, pages 18756--18768, 2023{\natexlab{b}}.
\newblock \doi{10.52202/075280-0822}.
\newblock URL \url{https://arxiv.org/abs/2305.10345}.

\bibitem[Yu et~al.(2022)Yu, Tang, Rao, Huang, Zhou, and Lu]{yu2022pointbert}
Xumin Yu, Lulu Tang, Yongming Rao, Tiejun Huang, Jie Zhou, and Jiwen Lu.
\newblock {Point-BERT}: Pre-training {3D} point cloud transformers with masked
  point modeling.
\newblock In \emph{Proceedings of the IEEE/CVF Conference on Computer Vision
  and Pattern Recognition (CVPR)}, pages 19313--19322, June 2022.
\newblock URL \url{https://arxiv.org/abs/2111.14819}.

\bibitem[Zhang et~al.(2026)Zhang, Yin, Yao, Qin, Zhang, Yang, Sun, Wang, Fan,
  Magno, and Zhang]{zhang2026mmmind}
Duo Zhang, Zhehui Yin, Zhiyun Yao, Haotong Qin, Xusheng Zhang, Hongliu Yang,
  Jianyu Sun, Junzhe Wang, Zizhou Fan, Michele Magno, and Daqing Zhang.
\newblock Teaching foundation models to read {mmWave}: Pose-guided kinematic
  representation for human behavior understanding.
\newblock \emph{arXiv preprint arXiv:2608.04127}, 2026.
\newblock URL \url{https://arxiv.org/abs/2608.04127}.

\bibitem[Zhang et~al.(2022)Zhang, Guo, Gao, Fang, Zhao, Wang, Qiao, and
  Li]{zhang2022pointm2ae}
Renrui Zhang, Ziyu Guo, Peng Gao, Rongyao Fang, Bin Zhao, Dong Wang, Yu~Qiao,
  and Hongsheng Li.
\newblock {Point-M2AE}: Multi-scale masked autoencoders for hierarchical point
  cloud pre-training.
\newblock In \emph{Advances in Neural Information Processing Systems},
  volume~35, pages 27061--27074, 2022.
\newblock \doi{10.52202/068431-1962}.
\newblock URL \url{https://arxiv.org/abs/2205.14401}.

\bibitem[Zhang et~al.(2024)Zhang, Xue, Wang, Geng, and Gao]{zhang2024radardiff}
Ruibin Zhang, Donglai Xue, Yuhan Wang, Ruixu Geng, and Fei Gao.
\newblock Towards dense and accurate radar perception via efficient cross-modal
  diffusion model.
\newblock \emph{IEEE Robotics and Automation Letters}, 9\penalty0 (9):\penalty0
  7429--7436, 2024.
\newblock \doi{10.1109/LRA.2024.3426389}.

\bibitem[Zhao et~al.(2021)Zhao, Jiang, Jia, Torr, and
  Koltun]{zhao2021pointtransformer}
Hengshuang Zhao, Li~Jiang, Jiaya Jia, Philip H.~S. Torr, and Vladlen Koltun.
\newblock Point transformer.
\newblock In \emph{Proceedings of the IEEE/CVF International Conference on
  Computer Vision}, pages 16259--16268, 2021.

\end{thebibliography}
\bibliographystyle{plainnat}

\appendix
\clearpage
\section{Extended Related Work}
\label{app:extended-related-work}

\textbf{General Point Cloud Pretraining.}
Self-supervised point cloud pretraining learns transferable representations from unlabeled data through contrastive learning, masked modeling, autoregressive generation, and self-distillation.
Contrastive methods relate features across different views of an observation.
PointContrast~\citep{xie2020pointcontrast} studies point-level alignment, while SegContrast~\citep{nunes2022segcontrast} studies segment-level alignment.
Masked Scene Contrast~\citep{wu2023msc} combines scene augmentation with masking.
PSA-SSL~\citep{nisar2025psassl} adds geometric prediction to retain object pose and size information.

Masked modeling learns local structure and context by predicting hidden regions.
Point-BERT~\citep{yu2022pointbert} uses a pretrained tokenizer to create discrete targets.
It predicts the tokens of masked point patches.
Point-MAE~\citep{pang2022pointmae} directly reconstructs masked patch coordinates.
Point-M2AE~\citep{zhang2022pointm2ae} uses a hierarchical encoder and decoder with masks consistent across scales.
It captures both local details and global structure.

For large-scale LiDAR scenes, GD-MAE~\citep{yang2023gdmae} uses a generative decoder to recover masked geometry.
GeoMAE~\citep{tian2023geomae} predicts local geometric attributes.
Occupancy-MAE~\citep{min2024occupancymae} uses range-aware voxel masking.
MAELi~\citep{krispel2024maeli} distinguishes visible and occluded space.
NOMAE~\citep{abdelsamad2025nomae} reconstructs occupancy in multiscale neighborhoods.
PointGPT~\citep{chen2023pointgpt} orders point patches by spatial proximity.
It predicts later patches from earlier ones to learn dependencies among spatial structures.
Sonata~\citep{wu2025sonata} applies self-distillation to a point cloud Transformer.
It enforces consistency across augmented views to reduce reliance on low-level geometric cues.
It also shows downstream transfer with a frozen encoder and a linear probe.



\textbf{Radar and Wireless Sensing Pretraining.}
Existing mmWave human sensing work learns transferable representations through self-supervised reconstruction and multi-source data integration.
These studies differ in input format, supervision, and downstream goals.

Wave2Body~\citep{liang2026wave2body} learns a radar tokenizer through masked reconstruction without pose labels.
It separately pretrains a body tokenizer on pose data.
Paired radar--pose data then train a mapping between the two tokenizers.
The full pipeline targets human pose estimation.
MAEPose~\citep{wei2026maepose} reconstructs masked spatiotemporal regions in range--Doppler videos.
It transfers the encoder to pose estimation.
Both studies mainly evaluate pretraining for pose recovery.
They do not systematically test one representation across diverse human sensing tasks.

WiSwiss~\citep{gong2026wiswiss} combines semantic deduplication, masked reconstruction, and transformation consistency across multiple wireless data sources.
It learns separate transferable backbones for WiFi and mmWave and evaluates downstream fine-tuning.
However, its public semantic-deduplication encoder uses class labels to select negative pairs for contrastive learning.
The full pipeline therefore depends on labels at this stage.
Its later backbone training uses self-supervised reconstruction and transformation-consistency objectives.
The mmWave input is a time--range--angle tensor.
Our work learns local states and their temporal evolution from sparse point clouds.

\textbf{Multimodal Fusion and External Supervision.}
X-Fi~\citep{chen2025xfi} uses modality encoders and dynamic fusion to support single modalities and their combinations.
Its feature extractors receive supervised training for each dataset and task.
It mainly addresses sensing under changing modality combinations.
WiCompass~\citep{liang2026wicompass} learns a pose space from motion-capture data.
It uses pose labels for radar samples to identify coverage gaps and guide later collection or synthesis.
It focuses on data coverage for pose estimation.
RadarLLM~\citep{lai2026radarllm} uses paired motion features to guide a radar tokenizer.
It then uses synthetic radar--text pairs for language alignment.
Its training requires additional motion and text information.
mmMind~\citep{zhang2026mmmind} first learns kinematic representations under synchronized 3D pose supervision.
It then performs radar--language alignment and instruction tuning.
Its encoder pretraining needs pose labels, even though inference uses only radar.


Dependence on shared task labels or cross-modal pairs limits the available data when a framework expands to more devices and sources.
Self-supervised pretraining can use radar sequences without these annotations.
Each downstream task can then train a prediction head with its own labels.
This lowers the annotation barrier for adding new data sources to pretraining.
We focus on multi-source mmWave human point clouds.
Pretraining uses only radar observations to learn shared prototype compositions and their temporal evolution.
It uses no action, pose, mesh, or text supervision.
We evaluate the same frozen encoder on data sources excluded from pretraining.
The tasks cover action classification, dense point generation, joint pose estimation, and future motion prediction.

\section{Dataset Description}
\label{app:dataset}

We expand the dataset overview in Table~\ref{tab:datasets} with details of the sensing streams, annotations, and acquisition settings.
Frame counts follow the summaries in the table, with captured frames distinguished from temporally standardized frame positions.
We identify sensors used only for annotation separately and distinguish hardware operating bands from acquisition bandwidths where the latter are reported.

\textbf{M4Human.}
M4Human provides four synchronized streams: RGB, depth, radar tensors, and radar point clouds.
It contains 20 subjects, 50 actions, 999 sequences, and about 661{,}000 frames.
The paper does not give a separate count of acquisition scenes.
Data were captured with a Vayyar vTrigB radar and an Intel RealSense D435 camera; a Vicon motion-capture system provided body annotations.
The radar operates over 62--69\,GHz with a 7\,GHz bandwidth at 12\,Hz.

\textbf{MARS.}
MARS pairs mmWave point clouds with Kinect skeleton measurements.
It covers four subjects and ten rehabilitation movements, with 80 sequences and 40{,}083 labeled frames (approximately 40K in Table~\ref{tab:datasets}).
An acquisition-scene count is not given.
Collection used a TI IWR1443BOOST radar and a Kinect V2 camera.
The radar starts at 77\,GHz, sweeps 3.2\,GHz, and records at 10\,Hz with 96 chirps per frame.

\textbf{milliFlow.}
milliFlow provides three synchronized streams: mmWave point clouds, RGB, and depth.
It covers 12 subjects, eight motion conditions, and three acquisition sites, with 43{,}200 frames after each recorded sequence is cropped to 200 frames.
Data were captured with a Vayyar vTrigB radar and an Intel RealSense D455 camera.
The radar uses stepped-frequency continuous waves over 62--63.6\,GHz with a 1.6\,GHz bandwidth, 20 transmit antennas, and 16 receive antennas.
The average collection rate is about 13.2\,Hz.

\textbf{MiliPoint.}
MiliPoint releases one sensing modality, mmWave point clouds; a ZED 2 stereo camera supplied skeleton labels.
It includes 11 subjects, 49 actions, and 545{,}059 frames.
The source does not enumerate distinct acquisition scenes.
The radar is a TI IWR1843 with three transmit and four receive antennas.
It operates over 77--81\,GHz with a 4\,GHz bandwidth and records at 24\,Hz.

\textbf{DGUHA.}
DGUHA (Dongguk University Human Activity) pairs mmWave point cloud sequences with Kinect skeleton annotations.
It covers 19 subjects and seven actions.
Each sequence is padded by repeating the last frame or truncated to 400 time steps.
Collection used a TI IWR1443BOOST radar and a Kinect sensor; the number of acquisition scenes is not specified.
The radar was operated at 20\,Hz with three transmit and four receive antennas; the capture chirp bandwidth is not reported.

\textbf{mmBody.}
mmBody provides three sensing streams: mmWave point clouds, RGB, and depth.
It contains 20 subjects and 100 motions across seven named scenes, with over 200{,}000 collected frames.
Collection used an Arbe Phoenix imaging radar, two Azure Kinect cameras, and an OptiTrack motion-capture system for body annotations.
The radar has 48 transmit and 48 receive channels and records at about 14\,Hz.
Its hardware covers 76--81\,GHz; the capture bandwidth of this dataset is not specified.

\textbf{mRI.}
mRI provides four measurement types: mmWave point clouds, RGB, depth, and inertial signals.
It includes 20 subjects, 12 action categories, 300 sequences, and approximately 160{,}000 synchronized frames, with 2D keypoint and 3D joint annotations.
The paper does not enumerate acquisition scenes.
The system uses a TI IWR1443BOOST radar, two Kinect V2 RGB-D cameras, and six wearable IMUs.
The radar starts at 77\,GHz, sweeps 3.2\,GHz, and records at 10\,Hz with 96 chirps per frame.

\textbf{MM-Fi.}
MM-Fi captures six data streams: RGB, infrared, depth, LiDAR, mmWave, and Wi-Fi CSI.
It covers 40 subjects and 27 actions in four environmental settings across two rooms, with 1{,}080 sequences and over 320{,}000 synchronized frames.
The release includes depth images and pose annotations; anonymized RGB images are provided separately, while image-derived keypoints represent the RGB and infrared streams in the standard data files.
Collection used an Intel RealSense D435, a TI IWR6843AOP radar, an Ouster OS1 LiDAR, and TP-Link N750 Wi-Fi devices.
The radar operates in the 60--64\,GHz band with three transmit and four receive antennas and supports up to 30 frames per second.
The released multimodal data are aligned at 10\,Hz; the exact capture chirp bandwidth is not specified.

\textbf{RadHAR.}
RadHAR releases one sensing modality: mmWave radar point clouds.
It contains two subjects and five actions, with approximately 167{,}000 frames based on the reported recording durations and frame rate.
The source does not report a distinct acquisition-scene count.
Collection used a TI IWR1443BOOST radar at 30\,Hz.
The device operates in the 76--81\,GHz band, while the exact capture chirp bandwidth is not documented.

\textbf{UniMM-HAR.}
UniMM-HAR is a derived radar-only benchmark built from RadHAR, mRI, and MM-Fi.
It combines 62 source subjects, 33 unified action categories, 40{,}494 sequences, and about 1.30 million normalized frame positions.
Each sequence is standardized to 32 frames through temporal resampling or zero padding; the resulting frame positions are not additional independent captures.

\par\noindent\begin{minipage}{\linewidth}
\section{Implementation Details}
\label{app:frame_adapter}

\subsection{Model Implementation}
\label{app:model_implementation}

PT$^3$ consists of a shared observation encoder, a prototype bank, and an autoregressive prediction module.
The shared observation encoder has five stages, in order: a Frame Adapter, point cloud encoder, device-conditioned FiLM, spatial readout, and shared FFN.
Table~\ref{tab:model_configuration} summarizes the model configuration, and Table~\ref{tab:pretrain_hyperparameters} lists the pretraining hyperparameters.

\medskip
\centering
\captionsetup{type=table}
\caption{Model configuration of PT$^3$.}
\label{tab:model_configuration}
\small
\renewcommand{\arraystretch}{0.95}
\setlength{\tabcolsep}{10pt}
\begin{tabular}{ll}
    \toprule
    \textbf{Parameters} & \textbf{Values} \\
    \midrule
    \multicolumn{2}{l}{\textbf{Frame Adapter}} \\
    \quad Frames per observation group & 3 \\
    \quad Maximum representatives per frame & 128 \\
    \quad MLP dimensions & $7\rightarrow32\rightarrow32$ \\
    \quad Output feature dimension & 39 \\
    \midrule
    \multicolumn{2}{l}{\textbf{Point Encoder}} \\
    \quad Encoder channels & $(36, 72, 144, 252, 504)$ \\
    \quad Encoder depths & $(2, 2, 2, 6, 2)$ \\
    \quad Voxel size & $0.01\,\mathrm{m}$ \\
    \quad Output projection & $504\rightarrow128$ \\
    \quad Normalization & LayerNorm \\
    \midrule
    \multicolumn{2}{l}{\textbf{Device-conditioned FiLM}} \\
    \quad MLP dimensions & $16\rightarrow32\rightarrow64\rightarrow128$ \\
    \quad Multiplicative gain range & $[0.95, 1.05]$ \\
    \quad Metadata dropout & 0.2 \\
    \midrule
    \multicolumn{2}{l}{\textbf{Spatial Readout and Shared FFN}} \\
    \quad Anchor grid & $3\times4$ in the $X$--$Z$ plane \\
    \quad Base Gaussian bandwidth $(x,y,z)$ & $(0.5, 1.0, 0.4)\,\mathrm{m}$ \\
    \quad Gaussian scale multipliers & $(0.75, 1.0, 1.25, 1.5)$ \\
    \quad Routing temperature & 0.25 \\
    \quad Null-route logit & $-4.5$ \\
    \quad Output size (slots $\times$ dimension) & $12\times128$ \\
    \quad Post-readout FFN dimensions & $128\rightarrow512\rightarrow128$ \\
    \midrule
    \multicolumn{2}{l}{\textbf{Prototype Bank}} \\
    \quad Bank size (prototypes $\times$ dimension) & $64\times128$ \\
    \quad Selected prototypes per slot & 4 \\
    \quad Assignment temperature & 0.08 \\
    \midrule
    \multicolumn{2}{l}{\textbf{Autoregressive Module}} \\
    \quad Transformer layers & 12 \\
    \quad Hidden dimension & 128 \\
    \quad Attention heads & 4 \\
    \quad Prediction MLP dimensions & $128\rightarrow128\rightarrow128$ \\
    \quad Prediction head output dimension & 64 \\
    \quad Temporal positional encoding & Group-level RoPE \\
    \bottomrule
\end{tabular}
\end{minipage}

\bigskip
\noindent\begin{minipage}{\linewidth}
\subsection{Hyperparameters}
\label{app:hyperparameters}

    \centering
    \captionsetup{type=table}
    \caption{Hyperparameters of pre-training.}
    \label{tab:pretrain_hyperparameters}
    \small
    \renewcommand{\arraystretch}{0.96}
    \setlength{\tabcolsep}{10pt}
    \begin{tabular}{ll}
        \toprule
        \textbf{Hyperparameters} & \textbf{Values} \\
        \midrule
        Epochs & 100 \\
        Batch size & 32 \\
        Optimizer & AdamW \\
        Learning rate (encoder and AR) & $1\times10^{-4}$ \\
        Learning rate (prototype bank) & $1\times10^{-4}$ \\
        AdamW $(\beta_1,\beta_2)$ & $(0.9, 0.999)$ \\
        AdamW $\epsilon$ & $1\times10^{-8}$ \\
        Weight decay (encoder and AR) & 0.05 \\
        Weight decay (prototype bank) & 0.01 \\
        Scheduler & Linear warmup + cosine decay \\
        Warmup epochs & 5 \\
        Minimum learning rate & $1\times10^{-5}$ \\
        Gradient clipping norm (both optimizers) & 1.0 \\
        Initial EMA momentum & 0.996 \\
        AR memory length & 32 observation groups \\
        Consistency groups per sequence & 4 \\
        Prototype loss weight ($\lambda_1$) & 1 \\
        Maximum consistency loss weight & 0.519 \\
        Consistency ramp-up epochs & 2 \\
        Calibration groups & 1024 \\
        Calibration standard-deviation floor & 0.01 \\
        Mixed precision & BF16 \\
        Random seed & 2026 \\
        \bottomrule
    \end{tabular}
\end{minipage}

\section{Downstream Evaluation Metrics}
\label{app:metrics}

We use task-specific metrics to evaluate the four downstream tasks. The reported measures are:

\begin{itemize}[leftmargin=*, itemsep=2pt, topsep=3pt]
\item \textbf{Accuracy} is the percentage of test samples for action classification whose predicted class matches the ground-truth class. We report it for both UniMM-HAR protocols and for DGUHA.
\item \textbf{Weighted F1} is the mean of the per-class F1 scores weighted by the number of test samples in each class. We report it for DGUHA action classification.
\item \textbf{Chamfer Distance (CD)} measures the geometric discrepancy between a generated point cloud and its depth-camera target in dense point generation. It averages squared Euclidean distances from each point to its nearest point in the other cloud in both directions, without requiring known point correspondences.
\item \textbf{Earth Mover's Distance (EMD)} measures point-cloud discrepancy through a one-to-one matching of the generated and target points, using squared Euclidean matching costs. Unlike CD, it accounts for how the full set of points is distributed. Both point-cloud metrics are computed after centering and displayed as 100 times their raw $\mathrm{m}^2$ values.
\item \textbf{Mean Per Joint Position Error (MPJPE)} is the mean Euclidean distance between predicted and ground-truth 3D joint positions, averaged over joints and test frames without per-pose alignment. We report it in millimeters for joint pose estimation.
\item \textbf{Procrustes-Aligned MPJPE (PA-MPJPE)} applies the same joint-position error after aligning each predicted pose to its ground truth by translation, rotation, and scale. It measures pose-shape error after removing differences in global placement and orientation, and is also reported in millimeters.
\item \textbf{Average Displacement Error (ADE)} measures future motion prediction over all ten predicted frames. For each frame, the error is the Euclidean norm of the concatenated XYZ differences for the 16 non-root joints, expressed in meters; ADE averages these frame-wise errors over the prediction horizon. In our evaluation, predicted joint centroids are aligned to the targets by translation only.
\item \textbf{Final Displacement Error (FDE)} uses the same future-pose error at the tenth and final predicted frame. For each test window, ADE and FDE are minimized independently over ten generated futures.
\end{itemize}

\section{Pretraining Baselines}
\label{app:baselines}

We compare PT$^3$ with five general point cloud pretraining methods and one mmWave-specific method.
Point-MAE, Point-BERT, Point-M2AE, and PointGPT-S are pretrained on the same unlabeled data as PT$^3$.
For downstream evaluation, we freeze each pretrained encoder and use the task-specific heads described in the following sections.

\textbf{Point-MAE}~\citep{pang2022pointmae}.
Point-MAE divides a point cloud into irregular local patches and masks a large proportion of them.
An asymmetric Transformer autoencoder reconstructs the masked patches from the visible ones.

\textbf{Point-BERT}~\citep{yu2022pointbert}.
A separately trained dVAE tokenizer assigns discrete targets to local point patches.
Point-BERT then masks input patches and trains a Transformer to predict their original tokens.

\textbf{Point-M2AE}~\citep{zhang2022pointm2ae}.
Point-M2AE uses pyramid-shaped encoder and decoder networks to learn point cloud features at multiple scales.
Its masking strategy keeps visible regions consistent across scales, while skip connections support hierarchical reconstruction.

\textbf{PointGPT-S}~\citep{chen2023pointgpt}.
PointGPT orders point patches by spatial proximity and predicts the next patch autoregressively.
Its extractor-generator Transformer uses dual masking; we adopt the smaller PointGPT-S configuration.

\textbf{Sonata}~\citep{wu2025sonata}.
Sonata learns point representations through self-distillation with a Point Transformer V3 encoder.
We use its publicly released pretrained weights.

\textbf{Wave2Body}~\citep{liang2026wave2body}.
Wave2Body pretrains an mmWave patch tokenizer through self-supervised reconstruction of masked geometry and radar attributes.
The tokenizer outputs continuous radar tokens.









\section{Action Classification}
\label{app:har}

This section describes the dataset, evaluation protocols, and implementation details for action classification.

\subsection{Dataset and Protocols}
\label{app:har_data}

We conduct action classification experiments on UniMM-HAR, which unifies mmWave point clouds from RadHAR, mRI, and MM-Fi into 33 action categories and provides two evaluation protocols: cross-subject (CSub) and cross-scene (CSet).

\begin{table}[h]
\centering
\caption{Summary of the two UniMM-HAR evaluation protocols. \#Act. denotes the number of action categories, while \#Train, \#Test, and \#Total denote the numbers of training, test, and total samples, respectively.}
\label{tab:har_protocols}
\small
\begin{tabular}{l l l c c c c}
\toprule
\textbf{Protocol} & \textbf{Data Sources} & \textbf{Split Criterion} & \textbf{\#Act.} & \textbf{\#Train} & \textbf{\#Test} & \textbf{\#Total} \\
\midrule
C-Sub & mRI, MM-Fi & Subject split (5:5) & 25 & 10{,}053 & 9{,}604 & 19{,}657 \\
C-Set & RadHAR, mRI, MM-Fi & Scene split (4:2) & 27 & 19{,}603 & 8{,}437 & 28{,}040 \\
\bottomrule
\end{tabular}
\end{table}

The two protocols represent different partitioning views of the same source datasets and are therefore not combined when counting samples.
Windows generated from the same original sequence do not cross data splits.

For supervised comparisons, we report published results for RadHAR~\citep{singh2019radhar}, FastHAR~\citep{an2022fuse}, and DAP-Net~\citep{lin2026dap} under the same evaluation splits.

\subsection{Implementation}
\label{app:har_impl}

\textbf{Frozen feature extraction.}
(1) Point-MAE: we extract features from the normalized tokens of the final encoder layer without applying masking.
(2) Point-M2AE: we extract features from the tokens of the final encoder stage using its native evaluation path.
(3) PointGPT: we extract features from the normalized output of the final pretrained causal encoder layer, excluding the SOS token.
(4) Point-BERT: we use the pretrained CLS representation from the query encoder together with its pretrained \texttt{cls\_head}.
(5) Sonata: we extract point features using its frozen encoder initialized with publicly released pretrained weights.
(6) Wave2Body: we extract features from the valid tokens produced by \texttt{encode\_tokens}.
(7) PT$^3$: we use the learned slot representations from the final encoder output.

\textbf{Unified linear probe.}
All pretrained encoders are frozen throughout downstream evaluation.
For each action classification sample, features extracted from three consecutive frames are fused to form a single sample-level representation.
The resulting features are standardized using statistics computed only from the corresponding training split.
We use the same single-layer linear probe for all methods, defined as $\mathrm{logits}=xW+b$.
The probe is fitted to one-hot labels using Ridge regression with a closed-form solution, without iterative optimization.
The regularization strength is selected by two-fold environment-wise cross-validation within the official training set, after which the probe is refitted using the full training split.


\section{Dense Point Generation}
\label{app:densification}

This section describes the dataset, evaluation protocol, and implementation details for dense point generation.

\subsection{Dataset and Protocols}
\label{app:dense_data}

mmBody~\citep{chen2022mmbody} is a public multimodal dataset for human mesh reconstruction, synchronously collected using mmWave radar, depth cameras, and RGB cameras.
It provides mmWave point clouds, depth point clouds, and 3D human mesh annotations across multiple indoor scenarios and challenging sensing conditions, including poor lighting, rain, smoke, and occlusion.
Most test scenarios contain unseen subjects, enabling evaluation under both environmental and cross-subject.

We follow the same split as mmWaveFlow~\citep{su2026mmwaveflow}.
The training set contains \texttt{sequence\_0} to \texttt{sequence\_19}, with 39{,}892 frames.
The test set contains two sequences from each of \textit{lab1}, 
\textit{lab2}, \textit{furnished}, \textit{poor\_lighting}, 
and \textit{rain}, totaling 20{,}207 frames; the smoke and occlusion scenarios are excluded.

For supervised comparisons, we report published results for mmPoint~\citep{xie2023mmpoint}, RadarHD~\citep{prabhakara2023radarhd}, RadarDiff~\citep{zhang2024radardiff}, LiDiff~\citep{nunes2024lidiff}, Tiger~\citep{ren2024tiger}, and mmWaveFlow~\citep{su2026mmwaveflow} under the same evaluation split.

\subsection{Implementation}
\label{app:dense_impl}


\textbf{Frozen feature extraction.}
(1) Point-MAE: we extract features from the final encoder tokens.
(2) Point-M2AE: we extract features from the tokens of the final encoder stage.
(3) PointGPT: we extract features from the real patch tokens of the final layer, excluding the SOS token.
(4) Point-BERT: we use the pretrained CLS representation from the query encoder together with its classification head.
(5) Wave2Body: we extract features from the valid context tokens.
(6) PT$^3$: we use the learned slot representations from the final encoder output.

\textbf{Unified projection head.}
The extracted frozen representations are standardized using statistics computed from the training set.
To accommodate different feature dimensions, all methods use the same projection structure, consisting of a $\mathrm{Linear}(D,512)$ layer, GELU activation, and a $\mathrm{Linear}(512,4096)$ layer.
The projected feature is reshaped to $128\times4\times8$ and fed into the same residual convolutional decoder to reconstruct 512 unordered 3D points sampled from the corresponding depth-camera point cloud.

\textbf{Training and evaluation.}
The downstream decoder is trained for 25 epochs using AdamW with a batch size of 128 and an initial learning rate of $3\times10^{-4}$.
The learning rate is decayed with a cosine schedule, and the training objective is $\mathcal{L}=\mathrm{CD}+0.19\,\mathrm{EMD}$.
We maintain an exponential moving average of the decoder parameters and use the final EMA weights for evaluation.
CD and EMD are computed after centering the predicted and target point clouds.

\section{Joint Pose Estimation}
\label{app:pose}

This section describes the dataset, evaluation protocols, and implementation details for joint pose estimation.

\subsection{Dataset and Protocols}
\label{app:pose_data}

We conduct joint pose estimation experiments on MM-Fi~\citep{yang2023mmfi}, which contains 27 daily activities performed by 40 subjects across four environments and provides mmWave point clouds with registered 3D skeleton annotations of 17 keypoints.
We follow the three official protocols: Random, Cross-Subject, and Cross-Environment, whose test splits contain 33{,}132, 33{,}257, and 41{,}541 frames respectively.
The task regresses the 3D coordinates of the 17 keypoints for the current frame, and we report MPJPE and PA-MPJPE in mm, where lower values indicate better performance.

The supervised comparison results for PointTransformer~\citep{zhao2021pointtransformer}, DiffPose~\citep{gong2023diffpose}, and mmDiff~\citep{fan2024diffusion} are taken from the mmDiff paper.

\subsection{Implementation}
\label{app:pose_impl}

\textbf{Frozen feature extraction.}
We use the same frozen feature readout as in the other downstream evaluations, and directly use the frozen frame feature cache of each encoder.

\textbf{Unified prediction head.}
All pretrained encoders are frozen, and only a TCN head is trained on top of the extracted features.
The features are standardized, and the first layer projects each model's features to 128 dimensions, after which all methods share four residual causal TCN blocks and a linear coordinate output layer.

\textbf{Training and evaluation.}
All methods follow the same data split, cleaning rules, and training configuration as our best-performing setting.
The number of training epochs is selected by PA-MPJPE on the validation set, after which the head is retrained from scratch on the union of the training and validation splits, and we report MPJPE and PA-MPJPE on the test set.
The TCN is trained for 100 epochs using AdamW with a batch size of 1024 and an initial learning rate of $3\times10^{-4}$. The learning rate is decayed with a cosine schedule.

\section{Future Motion Prediction}
\label{app:motion}

This section describes the dataset, evaluation protocols, and implementation details for future motion prediction.

\subsection{Dataset and Protocols}
\label{app:motion_data}

We conduct future motion prediction experiments on MM-Fi~\citep{yang2023mmfi}, which contains 27 daily activities performed by 40 subjects across four environments.
We use the mmWave point clouds and their registered 3D human skeleton annotations.
We follow the cross-subject protocol used in prior radar-based motion prediction work.
Sequences are divided into fixed-length non-overlapping windows.

All methods observe the same eight historical radar frames and predict ten future frames of joint trajectories.
The prediction target consists of the 3D positions of 16 non-root joints, represented relative to the pelvis in meters.
We report mean-aligned Average Displacement Error (ADE) and Final Displacement Error (FDE), where lower values indicate better performance.

The supervised comparison results for SPGSN~\citep{li2022psgsn}, HumanMAC~\citep{chen2023humanmac}, and mmPred~\citep{fan2026mmpred} are taken from the mmPred paper under the same MM-Fi evaluation protocol.

\subsection{Implementation}
\label{app:motion_impl}

\textbf{Frozen feature extraction.}
We use the same frozen feature readout for each pretrained encoder as in the other downstream evaluations.

\textbf{Unified prediction head.}
Since future motion prediction requires generating multiple future frames, we use a TCN as the downstream decoder to aggregate temporal features and predict the conditions required by the official mmPred diffusion model.
The predicted conditions are then fed into the frozen diffusion model to generate the future 10-frame pose sequence.
The TCN consists of an input projection followed by four residual causal convolution blocks; only the input projection dimension varies across encoders, while the remaining architecture is shared.

\textbf{Training and evaluation.}
We first train the TCN using pose mean-squared error and select the best checkpoint according to validation ADE.
The trained TCN is then used to generate the conditions required by the frozen official DDPM for future motion prediction.
The TCN is further optimized with
$
\mathcal{L}=\mathcal{L}_{\mathrm{noise}}+0.1\mathcal{L}_{\mathrm{condition}},
$
where only the TCN parameters are updated.

\section{Additional Results}

\subsection{Action classification on DGUHA.}
We also evaluate PT$^3$ on DGUHA~\citep{lee2023dguha} using its provided split of 607 training and 161 test clips.
For each point-cloud input variant, we freeze the encoder and fit a linear probe with Ridge regression.
Table~\ref{tab:dguha_har} reports the results.
AHC achieves 94.41\% accuracy and 94.52\% weighted F1.

\begin{table}[ht]
\centering
\caption{Action classification on DGUHA with a frozen PT$^3$ encoder and a Ridge-regression linear probe. ZP, GN, and AHC are the dataset's three processed point-cloud inputs.}
\label{tab:dguha_har}
\small
\setlength{\tabcolsep}{10pt}
\begin{tabular}{lccc}
\toprule
\textbf{Input} & \textbf{Accuracy (\%)} & \textbf{Weighted F1 (\%)} & \textbf{Correct / Total} \\
\midrule
ZP  & 91.30 & 91.19 & 147 / 161 \\
GN  & 86.96 & 86.98 & 140 / 161 \\
AHC & 94.41 & 94.52 & 152 / 161 \\
\bottomrule
\end{tabular}
\end{table}
\FloatBarrier

\subsection{Hyperparameter Analysis}
\label{app:hyperparameter_analysis}

\begin{figure}[!htbp]
\centering
\includegraphics[width=\linewidth]{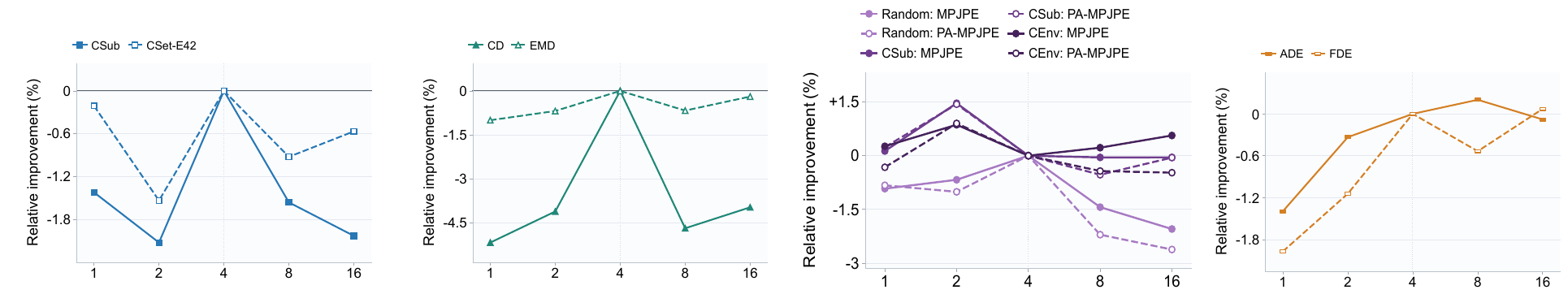}
\caption{Sensitivity to the number of selected prototypes per slot $k$.
The four panels show action classification, dense point generation, joint pose estimation, and future motion prediction from left to right.
The vertical axis gives relative improvement (\%) over the default $k=4$; positive values indicate better performance.}
\label{fig:hyperparam_topk}
\end{figure}

\begin{figure}[!htbp]
\centering
\includegraphics[width=\linewidth]{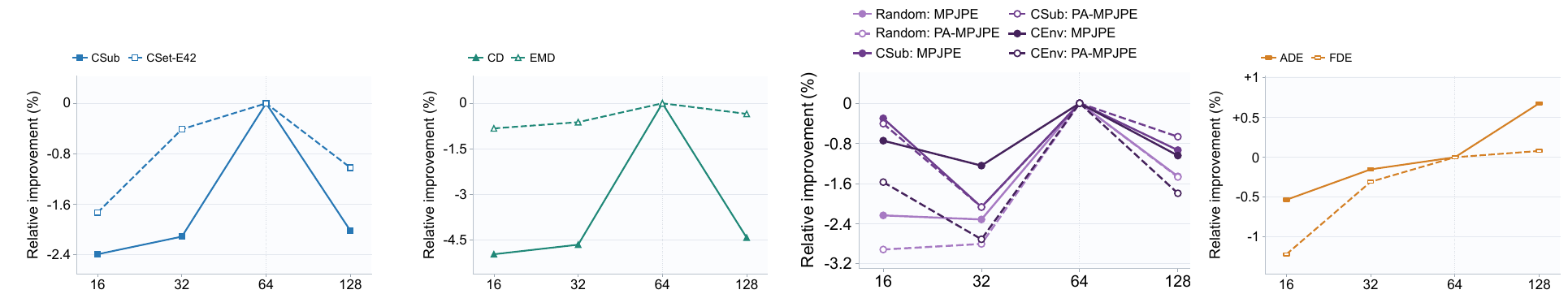}
\caption{Sensitivity to prototype bank size $K$.
The four panels show action classification, dense point generation, joint pose estimation, and future motion prediction from left to right.
The vertical axis gives relative improvement (\%) over the default $K=64$; positive values indicate better performance.}
\label{fig:hyperparam_prototypes}
\end{figure}

\begin{figure}[!htbp]
\centering
\includegraphics[width=\linewidth]{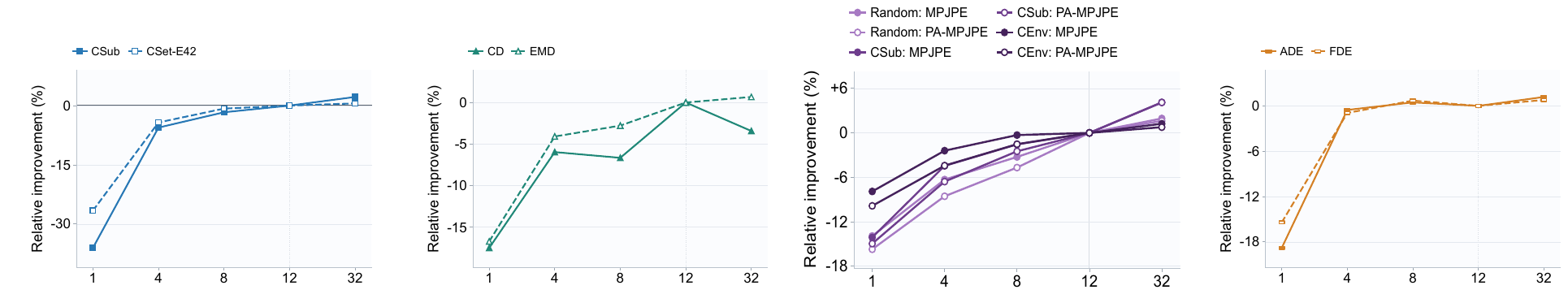}
\caption{Sensitivity to the number of spatial slots $S$.
The four panels show action classification, dense point generation, joint pose estimation, and future motion prediction from left to right.
The vertical axis gives relative improvement (\%) over the default $S=12$; positive values indicate better performance.}
\label{fig:hyperparam_slots}
\end{figure}

We vary the number of selected prototypes per slot $k$, the prototype bank size $K$, and the number of spatial slots $S$ one at a time.
Their default values are 4, 64, and 12, respectively (Table~\ref{tab:model_configuration}).
Figures~\ref{fig:hyperparam_topk}, \ref{fig:hyperparam_prototypes}, and \ref{fig:hyperparam_slots} show the relative performance changes.

\textbf{Top-$k$ Prototype Selection.}
Figure~\ref{fig:hyperparam_topk} shows that the default $k=4$ gives the most consistent results across tasks.
Reducing $k$ to 1 or 2 generally worsens action classification, dense point generation, and future motion prediction.
Increasing it to 8 or 16 yields no consistent gain.
Action classification and dense point generation usually decline, while joint pose estimation and future motion prediction results vary across protocols and metrics.
With too few prototypes, each slot may have limited capacity to compose local states.
With too many, the assignment may become less selective.

\textbf{Prototype Bank Size.}
Smaller prototype banks generally underperform $K=64$.
At $K=128$, future motion prediction improves slightly, but action classification, dense point generation, and joint pose estimation decline.

\textbf{Spatial Slot Count.}
Using only one or four spatial slots causes larger losses across tasks.
Increasing $S$ from 12 to 32 gives small, task-dependent changes.

\FloatBarrier


\subsection{Qualitative Action Prototype Heatmap}
\label{app:proto_heatmap}
\begin{figure}[h]
\centering
\includegraphics[width=0.8\linewidth]{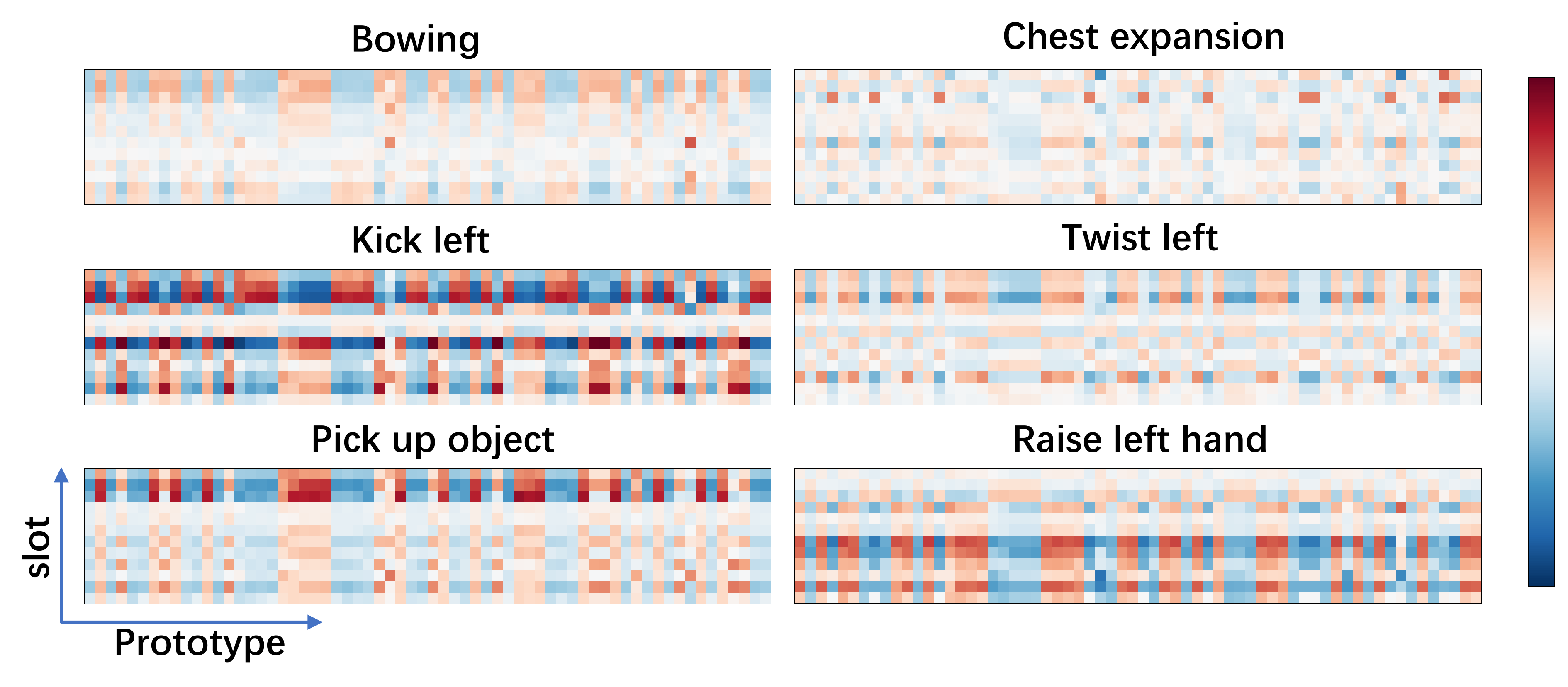}
\caption{Action prototype heatmap. Slot--prototype composition of one
representative subject on MM-Fi, averaged over the temporal windows of each
action and shown relative to the subject's mean assignment over the six actions.
Each panel corresponds to one action, with slots along one axis and prototypes
along the other.}
\label{fig:proto_motion}
\end{figure}

Figure~\ref{fig:proto_motion} shows the slot--prototype composition of six actions.
For each action we average the \(S\times K\) prototype assignment over the temporal windows.
The colour is relative to the mean assignment of this subject over the six actions.
Red means the action relies more on a prototype, and blue means it relies less.
White means the value is close to the subject's average.
Different actions show clearly different activation patterns.
Kick left and Pick up object form sharp bands along the slot axis.
Kick left has one strong band next to a suppressed band.
Pick up object concentrates its response on the upper slots and leaves a low band in the middle.
Raise left hand and Twist left are also banded, but weaker.
Bowing and Chest expansion are mostly white and deviate least from the average.
These per-action differences show that the combination of the shared prototype bank and the spatial slots is reorganised by the action state.

\subsection{Prototype Composition Detailed Results}
\label{app:proto_com}
\begin{figure}[h]
\centering
\begin{minipage}[t]{0.4\textwidth}
\centering
\includegraphics[width=\linewidth]{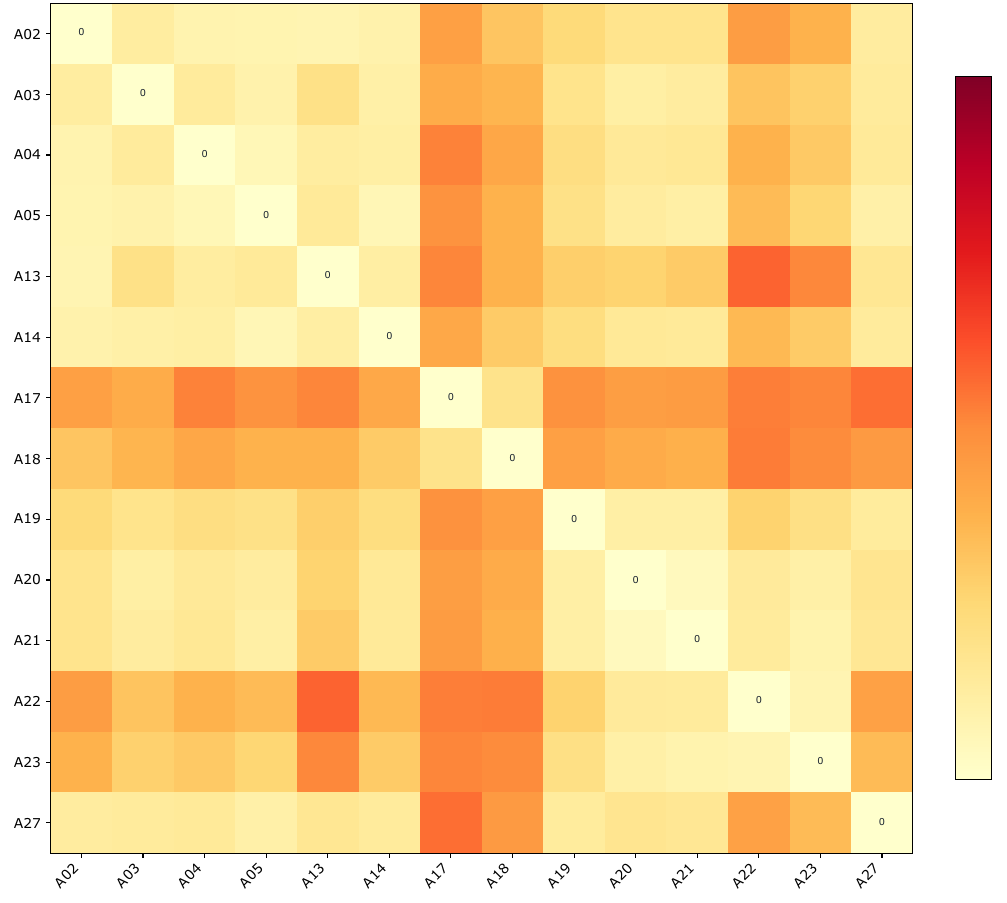}
\end{minipage}
\begin{minipage}[t]{0.4\textwidth}
\centering
\includegraphics[width=\linewidth]{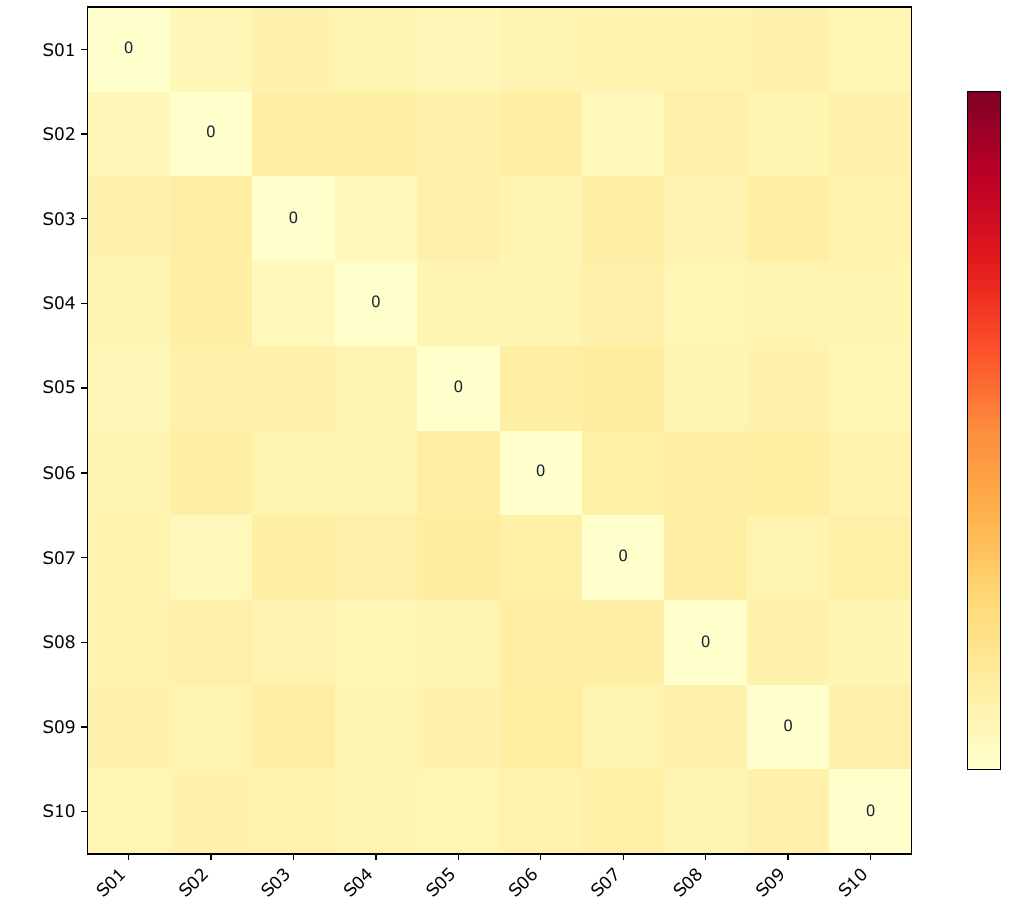}
\end{minipage}
\caption{Prototype-composition differences measured by Jensen--Shannon (JS) divergence.
Left: both axes list 14 action IDs, and each cell averages an action-pair distance over 10 subjects.
Right: both axes list 10 subject IDs, and each cell averages a subject-pair distance over 14 actions.
Both color bars encode JS divergence: pale yellow indicates smaller differences and dark red indicates larger differences.
Diagonal cells are zero because each item is compared with itself.}
\label{fig:proto_js_detail}
\end{figure}

Shared prototypes aim to represent actions across subjects through a common set of local state elements.
We test this idea with two comparisons: different actions within each subject and the same action across subjects.
These comparisons measure action sensitivity and cross-subject consistency, respectively.
For each sample, we average its \(S\times K\) prototype assignments over time and flatten the result.
We measure differences between the resulting compositions with Jensen--Shannon (JS) divergence.
Figure~\ref{fig:proto_js_detail} visualizes these two comparisons.
Action-pair distances are generally larger than subject-pair distances for the same action.
This pattern suggests action-dependent prototype usage across spatial slots.
Similar compositions across subjects support reusable action-related representations.

\section{Visualization}
\label{app:vis}

\begin{figure}[t]
\centering
\includegraphics[width=0.85\linewidth]{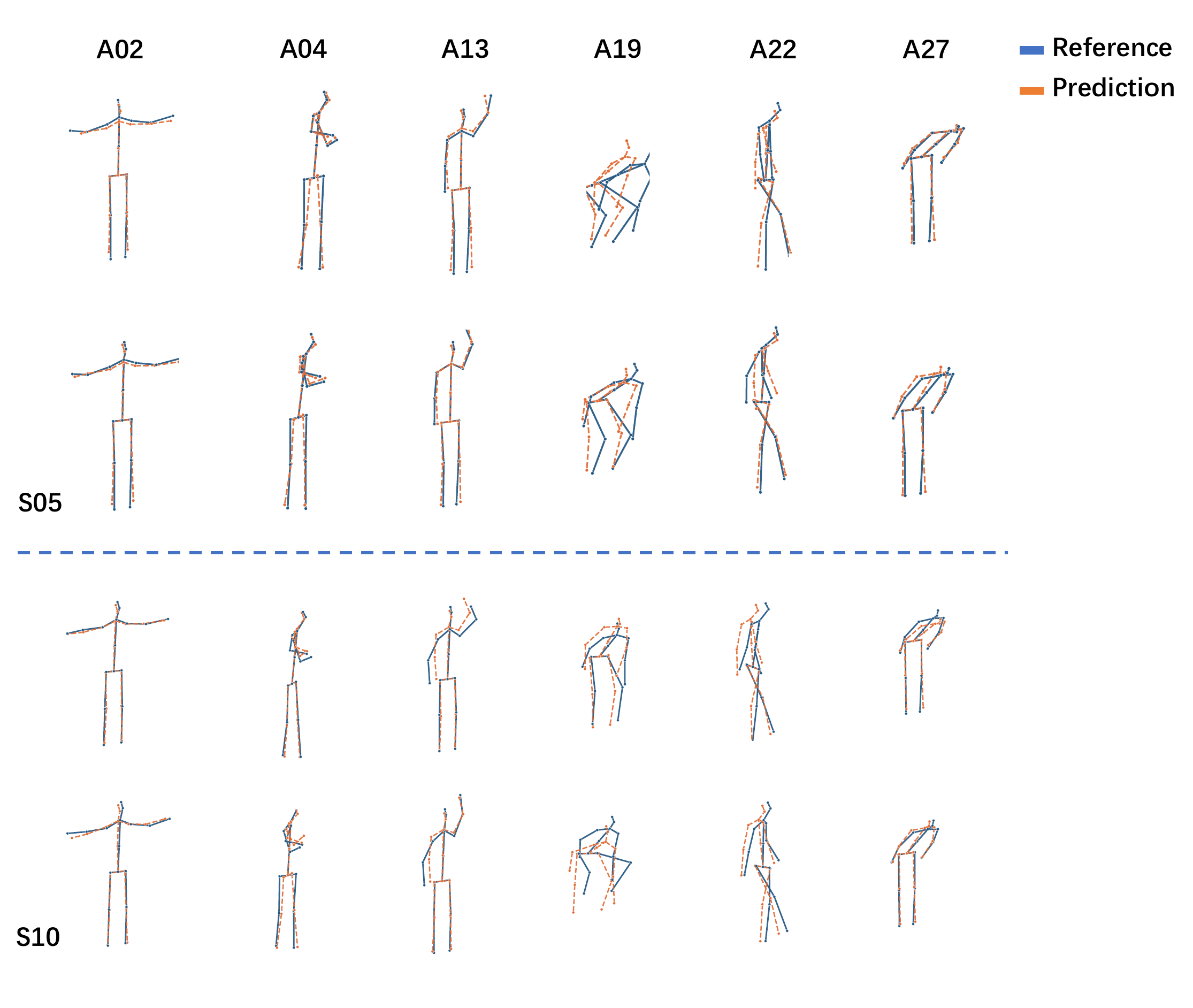}
\caption{Qualitative joint pose estimation results on six action categories:
A02 chest expand, A04 side twist to the left, A13 raise the left hand,
A19 pick up an object, A22 kick to the left, and A27 bow.}
\label{fig:pose_vis}
\end{figure}

Figure~\ref{fig:pose_vis} shows joint pose estimation results on six action categories.
We visualise two subjects.
Most predicted poses are accurate, while action A19 is less accurate.

\begin{figure}[h]
    \centering
    \includegraphics[width=0.9\linewidth]{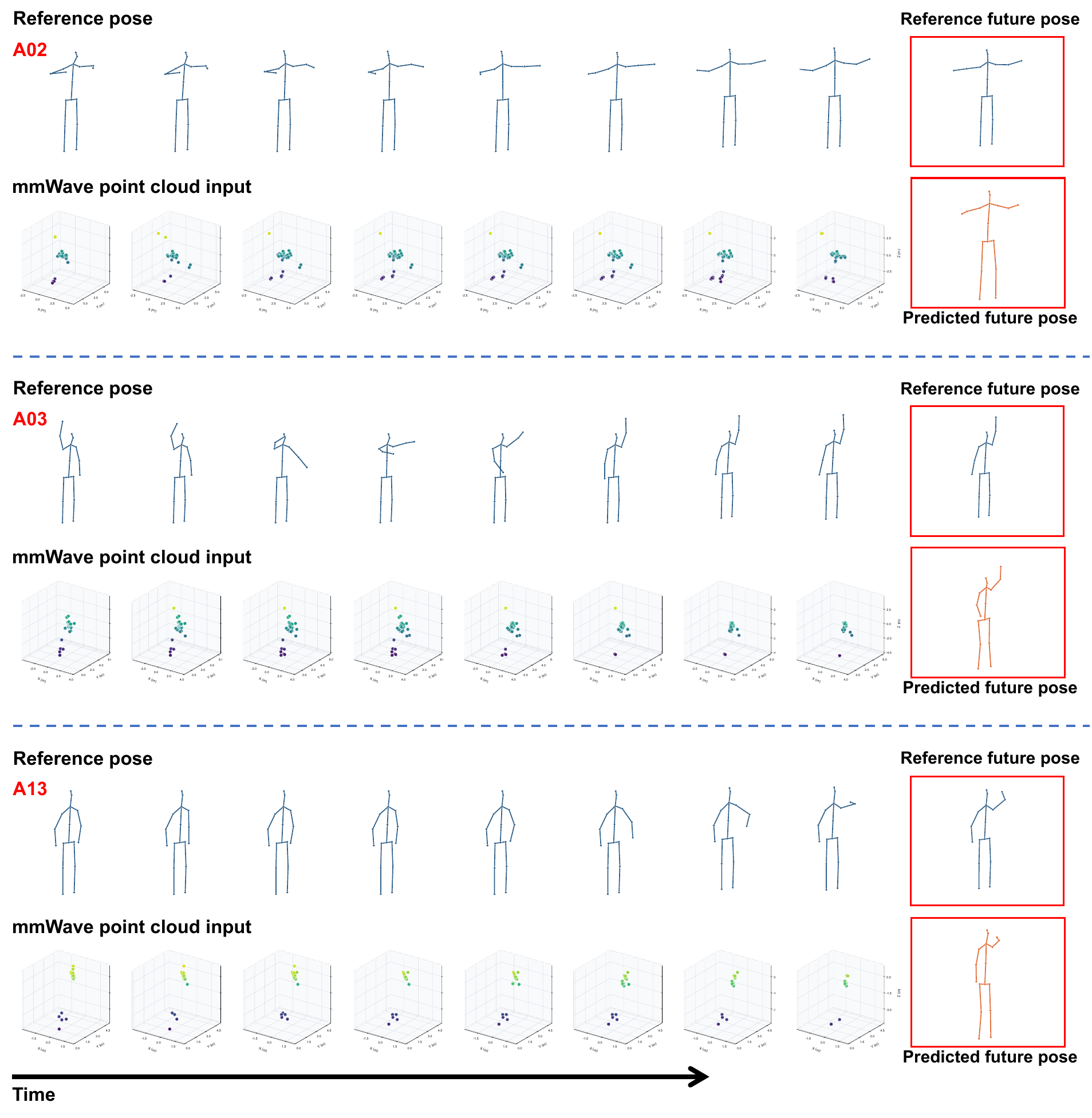}
    \caption{Qualitative future motion prediction results on actions A02, A03 and A13.
    From top to bottom: the input mmWave point cloud, the reference current pose,
    the predicted future pose, and the reference future pose.}
    \label{fig:pred_vis1}
\end{figure}

\begin{figure}[h]
    \centering
    \includegraphics[width=0.9\linewidth]{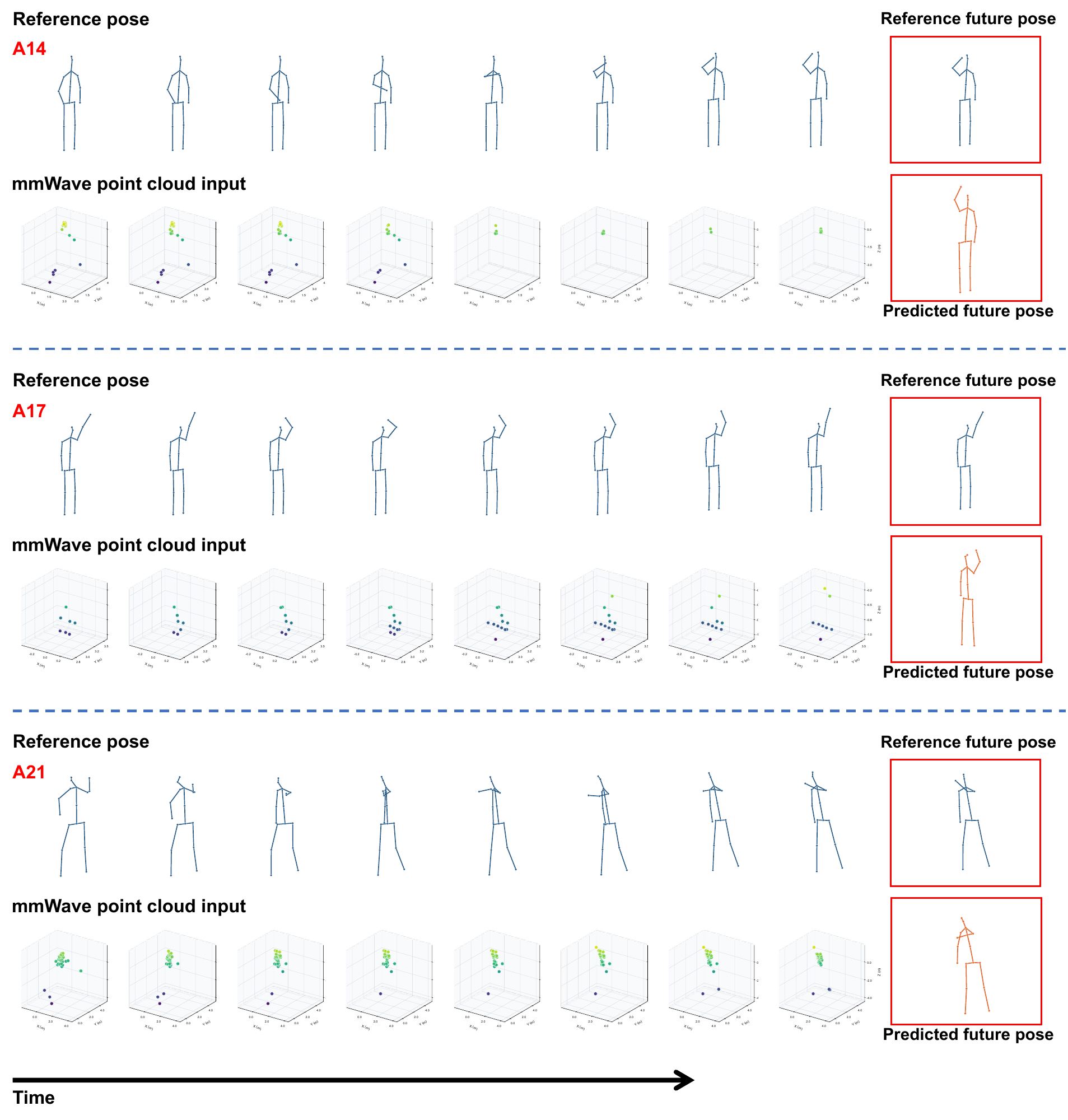}
    \caption{Qualitative future motion prediction results on actions A14, A17 and A21.}
    \label{fig:pred_vis2}
\end{figure}

Figures~\ref{fig:pred_vis1} and~\ref{fig:pred_vis2} show future motion prediction results on six action categories.
Each row contains the input mmWave point cloud, the reference current pose, the predicted future pose and the reference future pose.
The horizontal axis is time.

\section{Future Work}
\label{app:future_work}

(1) Efficient deployment of PT$^3$ on resource-constrained edge devices is a promising direction for future work.
Model compression, quantization, and lightweight implementations could reduce inference latency and memory consumption, supporting mmWave human sensing on low-power platforms.

(2) Developing radar-based world models of human motion is another promising direction.
Prototype compositions could serve as structured latent states for long-horizon dynamics modeling and conditional generation.
Such models could generate multiple plausible motion trajectories from sparse radar observations while characterizing uncertainty in future states.
This direction could provide dynamic priors for motion prediction and interaction planning in human--machine interaction and embodied intelligence.

\end{document}